\pdfoutput=1

\documentclass[11pt]{article}

\usepackage[final]{acl}
\usepackage{xcolor}
\usepackage{amsmath}
\usepackage{booktabs, float, pdflscape}
\usepackage{dblfloatfix}
\usepackage{adjustbox}
\usepackage{times, makecell}
\usepackage{latexsym, cleveref, multicol, caption}
\usepackage{multirow, makecell, cite,soul}
\usepackage[T1]{fontenc}
\newcolumntype{P}[1]{>{\centering\arraybackslash}p{#1}}
\usepackage[utf8]{inputenc}

\usepackage{xurl}     
\usepackage{hyperref} 
\usepackage{microtype}

\usepackage{inconsolata}
\usepackage{xcolor, subcaption}
\usepackage{tabularx}

\usepackage{graphicx}

\newcommand{\eat}[1]{}
\usepackage{authblk}

\title{A Generalizable Rhetorical Strategy Annotation Model Using LLM-based Debate Simulation and Labelling}

\author[1]{\textbf{Shiyu Ji}\thanks{Both authors contributed equally to this research.}}
\author[1]{\textbf{Farnoosh Hashemi$^*$}}
\author[1]{\textbf{Joice Chen}}
\author[1]{\textbf{Juanwen Pan}}
\author[2]{\authorcr\textbf{Weicheng Ma}}
\author[3]{\textbf{Hefan Zhang}}
\author[1]{\textbf{Sophia Pan}}
\author[3]{\textbf{Ming Cheng}}
\author[1]{\textbf{Shubham Mohole}}
\author[3]{\authorcr\textbf{Saeed Hassanpour}}
\author[3]{\textbf{Soroush Vosoughi}}
\author[1]{\textbf{Michael Macy}}
\affil[1]{Cornell University}
\affil[2]{Georgia Institute of Technology}
\affil[3]{Dartmouth College}
\affil[ ]{\textbf{Correspondence:} \texttt{\{sj787, sh2574\}@cornell.edu}}

\usepackage{balance}
\begin{document}
\maketitle
\begin{abstract}

Rhetorical strategies are central to persuasive communication, from political discourse and marketing to legal argumentation. However, analysis of rhetorical strategies has been limited by reliance on human annotation, which is costly, inconsistent, difficult to scale. Their associated datasets are often limited to specific topics and strategies, posing challenges for robust model development. We propose a novel framework that leverages large language models (LLMs) to automatically generate and label synthetic debate data based on a four-part rhetorical typology (causal, empirical, emotional, moral). We fine-tune transformer-based classifiers on this LLM-labeled dataset and validate its performance against human-labeled data on this dataset and on multiple external corpora. Our model achieves high performance and strong generalization across topical domains. We illustrate two applications with the fine-tuned model: (1) the improvement in persuasiveness prediction from incorporating rhetorical strategy labels, and (2) analyzing temporal and partisan shifts in rhetorical strategies in U.S. Presidential debates (1960–2020), revealing increased use of affective over cognitive argument in U.S. Presidential debates.
\end{abstract}
\section{Introduction}

Persuasion is a core mechanism in social influence  \citep{okeeffe2011persuasion}. It shapes how information is interpreted and acted upon across various domains, including marketing \citep{Kumar_Jha_Gupta_Aggarwal_Garg_Malyan_Bhardwaj_Ratn_Shah_Krishnamurthy_Chen_2023}, online communication \citep{anand_2011}, and political campaigns \citep{basave2016study}. In the political sphere, persuasion has become increasingly consequential amid rising polarization, growing partisan animosity, and widening ideological divides, with implications for democratic processes, public policy, and the sorting of partisan identities. \citep{druckman2022framework,iyengar2019origins,lelkes2016mass}  

Persuasion involves rhetorical strategies that engage either cognitive and affective processes  \citep{petty1986elaboration}. Cognitive arguments appeal to reason and evidence while affective arguments persuade by arousing emotional and moral reactions. These strategies are orthogonal to veracity. For example, empirical claims, even when fabricated, can lend credibility to misleading information \citep{serrano2021digital}, while emotional and moral appeals can go viral across social networks \citep{brady2017emotion,clifford2019emotional} and intensify affective polarization by provoking indignation and reinforcing group identities \citep{ding2023same}. 

The importance of rhetorical strategies in shaping consumer behavior, public discourse, and political polarization has attracted research utilizing datasets from online debates \citep{abbott-etal-2016-internet}, charity appeals \citep{wang-etal-2019-persuasion}, and commercial advertisements \citep{Kumar_Jha_Gupta_Aggarwal_Garg_Malyan_Bhardwaj_Ratn_Shah_Krishnamurthy_Chen_2023}. 

While prior studies offer valuable insights into persuasive techniques, the diversity of theoretical perspectives has led to inconsistent categorization of rhetorical strategies across human-annotated datasets. In addition, most existing datasets are focused on specific topical domains and rhetorical strategies, making it difficult to analyze the full range of persuasive techniques or generalize across domains \citep{Kumar_Jha_Gupta_Aggarwal_Garg_Malyan_Bhardwaj_Ratn_Shah_Krishnamurthy_Chen_2023}. These datasets often lack principled topic control, which obscures the distinction between rhetorical and topic-driven effects and leads models to overfit to topic-specific patterns with limited generalizability \citep{Chen_Yang_2021}. Most importantly, the cognitive and motivational demands of human annotation have resulted in a paucity of large-scale, high-quality datasets, and low inter-rater agreement compromises the establishment of reliable ground truth labels \citep{habernal_second}. These challenges have limited the development of robust deep-learning classifiers for automated identification of persuasive techniques.  

To address these challenges, we propose a novel framework that trains classifiers to detect rhetorical strategies using synthetic debate data generated and labeled by large language models (LLMs) and guided by a rhetorical typology informed by social and psychological theories of persuasion. Central to this framework is LLM labeling using simulated personas to annotate persuasive discourse. This automated annotation process enhances the reliability and scalability of rhetorical detection. 

Using this dataset, we trained a rhetorical classifier and validated the labels with human annotators. We then applied the classifier to analyze temporal trends in persuasive strategies in U. S. Presidential debates from 1960 to 2020.  Our analysis reveals shifting rhetorical patterns, providing new insights into the evolving landscape of partisan political communication. While our dataset generation focuses on the political domain, the framework is easily adaptable to other domains with minimal modification.

To sum up, our contribution is five-fold: 1) We present a \textbf{fully-automated scalable framework} for the generation and annotation of persuasive arguments that enhances the cross-context applicability of rhetorical labels. 2) We provide a \textbf{high-quality, topic-controlled dataset} that has been validated by human annotators.
3) We develop models to detect rhetorical strategies across varied topics and domains, with validation from human annotations and evaluation on external datasets. 4) Across five datasets from different domains, incorporating rhetorical labels into a fine-tuned BERT model improves performance in predicting persuasive outcomes, both within and across diverse datasets. 5) We identify a \textbf{significant increase in reliance on affective over cognitive strategies} during U.S. Presidential Debates going back to 1960, which may reflect the increase in affective polarization among both voters and political elites. 

\vspace{-0.25em}
\section{Related Work}
\vspace{-0.25em}
\subsection{Persuasion Strategy Identification}
Prior work labeling rhetorical strategies has relied on two sources: 1) persuasive arguments collected from existing corpora (e.g. college debates), and 2) crowd-sourced annotations \citep{wang-etal-2019-persuasion, habernal2016makes, habernal_second, Chen_Yang_2021}. These studies span multiple domains, including online conversations \citep{abbott-etal-2016-internet}, charity requests \citep{wang-etal-2019-persuasion}, commercial advertising \citep{Kumar_Jha_Gupta_Aggarwal_Garg_Malyan_Bhardwaj_Ratn_Shah_Krishnamurthy_Chen_2023}, and documented argumentation \citep{marro-etal-2022-graph}. Rhetorical labels are often derived from frameworks like Aristotle's typology of logos (logical reasoning), pathos (emotional appeals) and ethos (reference to credible sources) \citep{hidey-etal-2017-analyzing, rhetorical_strategies,stucki2018aristotelian}. For example, \citet{Higgins} annotated social environment reports for logos, pathos, and ethos. \citet{habernal_second} labeled 990 user-generated texts for logos and pathos, and \citet{abbott-etal-2016-internet} classified online discussion as emotion- or fact-based. Recent studies in computational linguistics have advanced automated rhetorical labeling by applying deep learning architectures to large annotated corpora. For example, \citet{yang-etal-2019-lets} developed a semi-supervised neural network model to classify persuasion tactics on social forums.  \citet{shaikh-etal-2020-examining} employed autoencoders (VAE) to analyze content and rhetorical strategies in loan requests. 

\vspace{-0.25em}
\subsection{Automatic Debate Generation}

The use of large language models (LLMs) in text generation has shown significant advantages across multiple applications, particularly in the social sciences where the ability to instantiate personas \citep{frisch2024llm, tseng2024two} is vital for nuanced and contextually appropriate outputs  \citep{veselovsky2023generating}. Even early LLMs like GPT-3 perform well at producing syntactically correct and semantically coherent text \citep{huang2024generating}, comparable to human-generated content \citep{munoz2024contrasting, dou2022gpt}, making them valuable tools for modeling social interactions and linguistic patterns \citep{xiao2023patterngpt}. LLMs are also effective for domain-specific tasks such as text generation for low-resource languages \citep{yang2024n}, where aligning with cultural and linguistic nuances is essential. 


\vspace{-0.25em}
\section{Rhetorical Strategies}
\label{sec:def}

We use a rhetorical typology that integrates Aristotle’s classical framework with the dual-process distinction between cognitive and affective persuasion \citep{petty1986elaboration, chaiken1999dual}. Reasoning with logic and evidence involves cognitive processes, while emotional and moral arguments are affective. 

Studies based on Aristotle's typology use logos inconsistently, sometimes referring to evidence and other times to logical reasoning \citep{egawa-etal-2019-annotating, iyer_2019, marro-etal-2022-graph}. This can be especially problematic in annotation tasks. Moreover, while logos refers to logical reasoning, the rules of formal logic are overly narrow and difficult to operationalize for annotation. We therefore separate reasoning and evidence into two distinct strategies and focus on \textit{causal }reasoning, in which an argument points to the positive or negative consequences of an action or event \citep{walton2012argument}. 

On the affective side, we distinguish between emotional and moral arguments. Appeals to emotion have been identified across multiple domains \citep{yang-etal-2019-lets, cabrio_proceedings_2018, abbott-etal-2016-internet} and involve the expression of evocative language to arouse emotions in the target audience \citep{miceli2006emotional}. Evocative language can also include moral emotions such as compassion, harm, betrayal, and degradation \citep{haidt2003moral, feinberg2019moral, anand_2011}. However, we classify these as moral persuasion, which is distinct from non-judgmental emotional appeals in that they refer to normative and ethical principles \citep{anand_2011, iyer_2019, yang-etal-2019-lets, feinberg2019moral}.

These distinctions yield the following four-fold typology (see \Cref{sec:strategytable} for examples and illustrations of each):

\textbf{Causal} -  \textit{A causal argument relies on cause-and-effect reasoning to explain or predict the positive or negative consequences of an action that are measurable or observable, with or without evidence.}
 
\textbf{Empirical} - \textit{An empirical argument relies on evidence such as statistics, examples, illustrations, anecdotes, and/or citations to sources that support the argument}.

\textbf{Emotional} -  \textit{An emotional argument relies on impassioned, arousing, or provocative language to express or evoke feelings (such as frustration, fear, hope, joy, desire, sadness, hurt, and/or surprise). }

\textbf{Moral - }\textit{A moral argument relies on concepts of right and wrong, justice, virtue, duty, or the greater good in order to persuade others about the ethical merit of a position, decision, or behavior.}

\begin{figure*}
    \centering
    \includegraphics[width=0.95\linewidth]{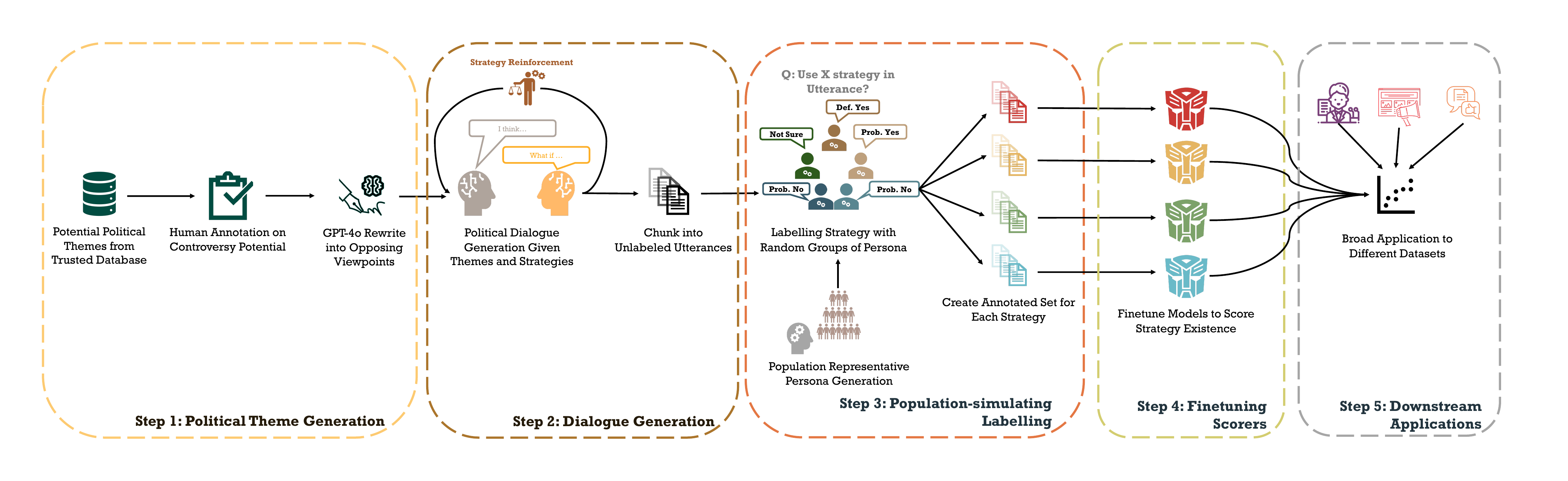}
    \caption{Overview of our proposed framework.}
    \label{fig:framework}
\end{figure*}

\vspace{-0.5ex}
\section{Methods}

Using this typology, we developed a machine classifier for automated labeling of rhetorical strategies. Our approach consists of the five steps illustrated in \Cref{fig:framework}: 1) identifying controversial political topics; 2) using LLMs to generate political debate dialogues;  3) prompting LLMs to annotate the generated dialogues;  4) fine-tuned a model for strategy classification;  5) applying the fine-tuned model to downstream analytical tasks.

\subsection{Opposing Stances Generation}
\label{sec:stance_gen}
To develop a persuasion strategy detection model for political texts, we used a combined human annotation and LLM keyword elaboration to generate diverse stances on controversial issues, ensuring balanced dialogues for robust model training. We identified controversial political topics in the United States using the Opposing Viewpoints database \emph{Opposing Viewpoint} provided by \emph{Gale} \citep{gale_opposing_viewpoints}, a trusted publisher of research content that offers diverse perspectives on contemporary social issues in the U.S. This yielded a list of 475 topical keywords (e.g., abortion, for-profit education, U.S. budget deficit). We used human annotation to refine the keyword list to those encompassing  opposing viewpoints. Two annotators were tasked with answering "yes" or "no" to this question: \emph{"Based on the provided keyword, are at least two distinct and opposing viewpoints evident in public discussions within the United States?"} Topics where both annotators answered "yes" were retained, resulting in a refined list of 146 contentious keywords.

Next, we used GPT-4o to expand each topic keyword into two broad opposing stances. These stances that were then used in our dialogue generation framework to create diverse and flexible argumentation, with broad topical coverage. This yielded 146 paired opposing arguments, associated with the 146 controversial topic keywords, which we used to generate debates. The prompt for generating paired opposing stances, along with examples, is provided in \cref{tab:topics_stances} in Appendix \ref{sec:topicprompt}. Each topic was labeled as political or nonpolitical by two independent
human annotators, with a third annotator resolving any disagreements, yielding 121 political and 25 non-political topics.


\subsection{Controlled Debate Generation with Topic and Rhetorical Strategy Constraints} 
\label{sec:dialogue}

We adapted the automated debate generation framework from \citet{Ma_et_al_2025} to simulate multi-turn english dialogues between two LLM agents, using the opposing stances generated from the 146 topics. Agents were prompted to either adopt or avoid one of four rhetorical strategies (causal, empirical, moral, or affective), ensuring a balanced distribution of strategies across topics and mitigating topic driven effects in downstream detection tasks.

For each generated argument from an agent in a debate turn, a detection agent evaluated whether it aligned with the assigned strategy and prompted revisions when necessary. This detect-and-revise process could occur up to two times per argument, improving the rhetorical fidelity of generated debates. Two additional agents are employed to enhance dialogue quality. One refines individual arguments to avoid redundancy and trivial language use, and the other oversees the integrity of the generation process after each round, ensuring logical consistency within each dialogue and determining when the dialogue should conclude. (Full agent instructions are reported in \Cref{sec:prompts_debate}.)  This process generated eight strategy-specific dialogues for each of the 146 controversial topics with a maximum five rounds of arguments, totaling 11,420 arguments, each with an average length of 63.4 words.  

\subsection{LLM-Based Persuasion Scoring}
\label{sec:scoring}
We used LLM  annotation to quantify the extent to which each rhetorical strategy—causal, emotional, empirical, and moral—was exhibited in model-generated arguments, we employed large language models (LLMs) as annotators. Recent studies have demonstrated that LLMs exhibit strong alignment with human judgment in multiple domains, including clinical text summarization \citep{clinical}, moral judgment \citep{moral}, sentiment classification, political leaning detection \citep{sentimentllm} and replicating human decision patterns in social dilemma experiments \citep{aher}. Prior research suggests that prompting the model with role-specific or identity-related persona, can enhance annotation quality by encouraging more consistent and contextualized responses~\citep{el-baff,simulation2, simulation3,simulation4,simulation5,simulation6}. Accordingly, we used five instances of GPT-4o to independently evaluate and score each argument, each from the standpoint of a different assigned persona. Each persona had a unique demographic profile based on sex, age, race, education, and partisan affiliation, with each profile aligned with the joint probabilities for the U.S. adult population, such that age, sex, and race were statistically independent while the correlations with education and political leaning reflected those in the underlying population, using data from the U.S. Census  \citep{usbureaudata2025}, American Council on Education  \citep{acedata2024}, and Pew Research Center  \citep{pewdata2024}. The five profiles increased variability across the LLM annotators and enhanced the interpretive diversity observed among human annotators.See Appendix \ref{sec:personaapp} for details on persona construction.

Each model, aside from its assigned persona, received the same prompt containing operational definitions of the four rhetorical strategies and two illustrations per strategy. Illustrations were drawn from Moral-Emotions \citep{kim-etal-2024-moral}, Ethixs \citep{vrakatselietal24-ethixs}, and UKPConvArg \citep{habernal2016makes}, and were included only if independently labeled with full agreement by three human annotators. The prompt asked the LLM to rate each argument on a five-point Likert scale (1 = definitely not using, 5 = definitely using, 3 = uncertain) for each strategy. This yielded four scores per argument, one per strategy. For each strategy, we further averaged across five persona-conditioned LLM annotations. For downstream training, scores were linearly mapped to a 0 to 1 scale using $(x - 1)/4$, where 0 = definitely not using, 1 = definitely using, and 0.5 = uncertain. The full prompt is included in \cref{sec:annptationprompt}.

\label{sec:gpt-annotated-score}
\begin{figure*}[t]
    \centering
\includegraphics[width=\textwidth]{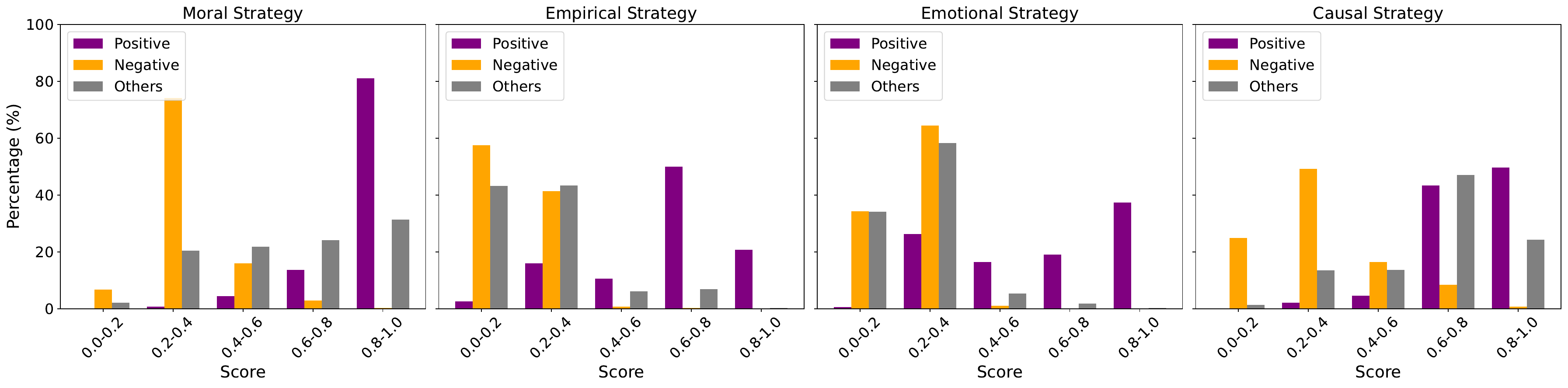}
    \caption{\textbf{Distribution of LLM-assigned strategy scores for utterances conditioned to use (Positive), avoid (Negative), or use a different rhetorical strategy (Others) for each target strategy.} Positive utterances were generated with prompts instructing the model to use the corresponding strategy; Negative utterances were prompted to avoid it; and Others includes utterances that were prompted for one of the other three strategies.}
    \label{fig:gptscores}
\end{figure*}


To evaluate the effectiveness of the rhetorical constraints described in Section~\ref{sec:dialogue}, we examined the distributions of LLM-based scores for arguments conditioned to use or avoid each strategy. As shown in Figure \ref{fig:gptscores} \eat{(Appendix \ref{sec:gpt-annotated-score})}, scores for all four strategies were consistently higher for positive (use) cases than for negative (avoid) ones across all four strategies. The Spearman correlations between the binary assignment (use vs. avoid) and the corresponding LLM-assigned strategy scores are reported in Table \ref{tab:spearman_gptlabels} \eat{(Appendix \ref{sec:gpt-annotated-score})}, showing strong associations for moral ($\rho = 0.863$), emotional ($\rho = 0.785$), causal ($\rho = 0.812$), and empirical ($\rho = 0.805$) strategies.

The datasets were used to fine-tune models dedicated for rhetorical strategy identification for downstream application. The results are shown in \Cref{sec:classifier}.

\vspace{-1ex}

\section{Results}
\vspace{-0.5ex}
\label{sec:results}
\subsection{Human Validation Study}

We validated the rhetorical strategy labels assigned to the LLM-generated debates through an annotation study conducted on Qualtrics, involving 355 college-educated english-speaking participants recruited via Prolific. Each participant annotated eight arguments randomly sampled from the LLM-generated debates in the test dataset of \Cref{sec:performance} used to evaluate the final model. They also annotated two other arguments from U.S. Presidential debates between 2000 and 2012, balanced for partisanship (to validate the downstream task in \Cref{subsec:presidential}). For each argument, participants were asked to rate the extent to which each of the four rhetorical strategies was present, using the same Likert scale as the LLM annotation. In total, 728 arguments from the LLM-generated debates and 182 from the Presidential debate corpus were evaluated. 

Prior to annotation, all participants completed a training session that explained the four rhetorical strategies, followed by a comprehension quiz to ensure that annotators understand the definition to ensure annotation quality.  To improve label reliability, we used arguments annotated by at least three annotators and took the average rating (mapped into 0 to 1 as according to \Cref{sec:scoring}) per argument. This yielded 587 arguments with human labels from synthetic debates and 147 from Presidential debates. Results of the study are reported in \Cref{subsec:human_valid} and \Cref{sec:performance}.

\vspace{-0.75em}
\subsubsection{Human Validation on LLM-generated Debate Quality and LLM-Scoring Quality}
\label{subsec:human_valid}

\begin{figure}[t]
    \centering
    \includegraphics[width=0.85\columnwidth]{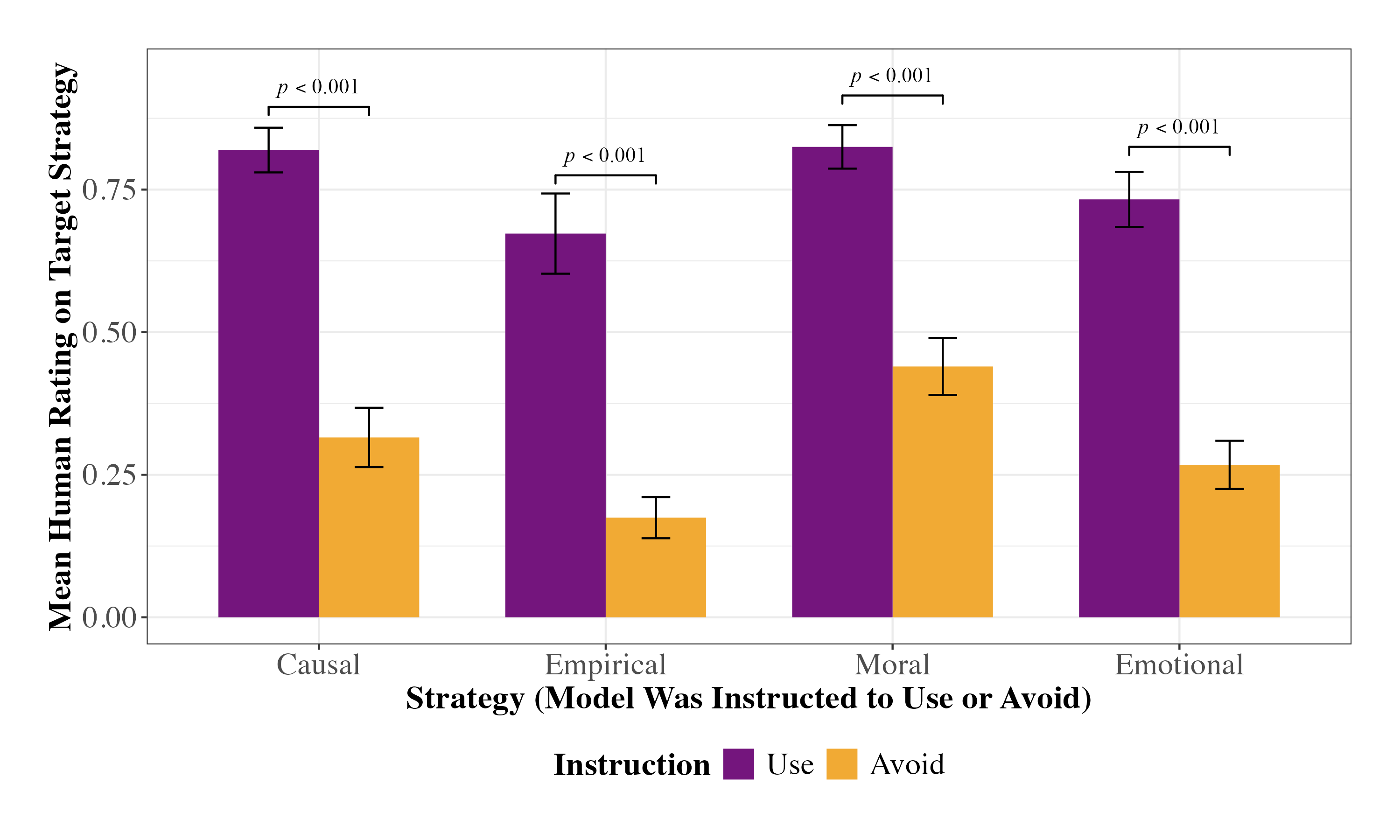}
    \caption{Human-labeled strategy scores for dialogues instructed to use vs. avoid each persuasion strategy. Scores range from 0 (definitely not using) to 1 (definitely using), with 0.5 indicating uncertainty.}
    \label{fig:strategy_gen_valid}
    
\end{figure}

\Cref{fig:strategy_gen_valid} reports the average human scores for the target strategy, depending on whether the LLM was instructed to use versus avoid that strategy, along with \textit{t}-tests for the difference between "use" and "avoid." All strategies show substantial and highly significant differences, demonstrating the effectiveness of our strategy-specific synthetic debate generation framework.

\begin{table}[t]
\centering
\resizebox{0.95\columnwidth}{!}{
\begin{tabular}{l|c|c|c|c}
\toprule
\textbf{} & \textbf{Moral} & \textbf{Emotional} & \textbf{Causal} & \textbf{Empirical} \\
\midrule
\textbf{\# of utterances} & 2848 & 2832 & 2862 & 2878 \\
\textbf{Spearman's $\rho$} & 0.863 & 0.785 & 0.812 & 0.805 \\
\bottomrule
\end{tabular}
}
\caption{Number of utterances and Spearman correlation for each rhetorical strategy (all results are significant, $p < 0.0001$}
\label{tab:spearman_gptlabels}
\end{table}

\begin{table*}[!t]
\centering
    \resizebox{\textwidth}{!}{
    \begin{tabular}{p{4cm}p{6cm}|P{3cm}P{3cm}|P{3cm}P{3cm}|P{3cm}P{3cm}}
    \hline
     &  & \multicolumn{2}{P{6cm}}{\textbf{In-Domain Evaluation}}& \multicolumn{2}{|P{6cm}|}{\textbf{Out-Distribution Evaluation}}&  \multicolumn{2}{P{6cm}}{\textbf{Cross-Domain Evaluation}}\\\cline{3-8}
    \textbf{Strategy} & \textbf{Pretrained Model for Fine-tuning} & \textbf{RMSE $\downarrow$} & \textbf{Spearman's} $\mathbf{\rho}$ $\uparrow$ & \textbf{RMSE $\downarrow$} & \textbf{Spearman's} $\mathbf{\rho}$ $\uparrow$& \textbf{RMSE $\downarrow$} & \textbf{Spearman's} $\mathbf{\rho}$ $\uparrow$ \\\hline
    Causal & ROBERTa-base & \textbf{0.099} (0.005) & \textbf{0.870} (0.000)& \textbf{0.102} (0.000)& \textbf{0.865} (0.002)& \textbf{0.116} (0.002)& \textbf{0.850} (0.002)\\
     & LLaMA-3.2-Instruct-3B + QLoRA & 0.110 (0.001)& 0.820 (0.007) & 0.109 (0.003) & 0.820 (0.015)& 0.118(0.005) & 0.808 (0.011)\\
    \hline
    Empirical & ROBERTa-base & \textbf{0.077} (0.004) & \textbf{0.931} (0.002) & \textbf{0.079} (0.003) & \textbf{0.922} (0.001) & \textbf{0.084} (0.002) & \textbf{0.913} (0.002)\\
     & LLaMA-3.2-Instruct-3B + QLoRA & 0.089 (0.003) & 0.911 (0.006) & 0.087 (0.003) & 0.899 (0.007)& 0.093 (0.002)& 0.903 (0.001)\\\hline
    Emotional & ROBERTa-base & \textbf{0.072} (0.002) & \textbf{0.872} (0.002)  & \textbf{0.073} (0.001) & \textbf{0.864} (0.002) & \textbf{0.082} (0.001) & \textbf{0.887} (0.001)\\
     & LLaMA-3.2-Instruct-3B + QLoRA & 0.083 (0.002) & 0.852 (0.008) & 0.079 (0.001)& 0.841 (0.005) & 0.091 (0.001) & 0.854 (0.012)\\\hline
     Moral & ROBERTa-base & 0.102 (0.005) & \textbf{0.939} (0.004)& 0.107 (0.004)& \textbf{0.935} (0.001) & 0.132 (0.004) & \textbf{0.915} (0.002) \\
     & LLaMA-3.2-Instruct-3B + QLoRA & \textbf{0.099} (0.003) & 0.932 (0.003) & \textbf{0.102} (0.003) & 0.932 (0.003) & \textbf{0.117} (0.002) & 0.910 (0.004) \\
    \hline
    \end{tabular}
    }
    \caption{Transfer Learning Performance on AI-Generated Debate Data. We fine-tuned each pretrained model three times per persuasion strategy and report the mean and standard deviation on the test sets. Performance was evaluated using Spearman correlation and RMSE against LLM-based scores. RoBERTa outperformed LLaMA and showed minimal performance drop in cross-domain tests with non-politcal topics (e.g., 0.024 for moral strategy).}

    \label{tab:transfer_learning}
\end{table*}

We validated our LLM-based persuasion scoring using the synthetic debate data from the human-annotated set. (Due to budget constraints, LLM scoring was not applied to the presidential debate data, though external corpora were used for additional validation; see \cref{sec:valid}.) The LLM scoring showed strong spearman correlations with human annotations for causal ($\rho = 0.612$), empirical ($\rho = 0.622$), emotional ($\rho = 0.599$), and moral strategies ($\rho = 0.716$), all significant at $p < 0.001$.

\vspace{-2ex}
\subsubsection{Reliability and Quality of LLM Versus Human Annotation}

While human annotation has long been the standard for creating ground-truth datasets, in the annotation study, we observed that large language models (LLMs) provide a more reliable and scalable alternative for rhetorical strategy annotation. We support this claim with the following three observations.

First, despite being theoretically motivated and providing richer information than binary classifications, human annotation of persuasion strategies is less reliable and requires more annotators per sample when fine-grained scales are used. In our study, we observed low inter-rater agreement (average Cohen’s $\kappa = 0.148$; see Table~\ref{tab:human_kappa}) among human annotators using the five-class scheme across all rhetorical strategies, while agreement improved under coarser schemes (average Cohen’s $\kappa = 0.281$ in a three-class setting, i.e., \emph{yes} vs. \emph{uncertain} vs. \emph{no}, and $\kappa = 0.321$ in a binary setting, i.e., \emph{yes} vs. \emph{no/uncertain}; see Table~\ref{tab:human_kappa}). This suggests that much of the disagreement stems from scale granularity rather than fundamental interpretive differences, which also necessitates our approach to aggregate annotations from at least three human annotators to establish a reliable ground truth. While feasible for our study, such aggregation is costly and limits scalability.

\begin{table}[t]
\centering
\small
\resizebox{\columnwidth}{!}{
\begin{tabular}{lccc}
\toprule
 &  \multicolumn{3}{c}{\textbf{Classification Scheme}} \\
\cline{2-4}
\textbf{Rhetorical Strategy} & \textbf{Five-Class} & \textbf{Three-Class} & \textbf{Two-Class} \\
&\textbf{(Original Scheme)}&& \\
\midrule
Causal     & 0.151 & 0.294 & 0.314 \\
Empirical  & 0.141 & 0.290 & 0.334 \\
Moral      & 0.146 & 0.287 & 0.324 \\
Emotional  & 0.153 & 0.251 & 0.312 \\
\midrule
\textbf{Average} & 0.148 & 0.281 & 0.321 \\
\bottomrule
\end{tabular}
}
\caption{Human inter-rater agreement (Cohen’s Kappa) across rhetorical strategies under different classification schemes. Agreement improves under coarser schemes, indicating that variability stems largely from scoring granularity.}
\label{tab:human_kappa}
\end{table}

Second, compared to individual human annotators, individual LLMs align more closely with aggregated human consensus. We assessed this by constructing a Leave-One-Out (LOO) Human Ground Truth, and comparing left-out human labels or LLM outputs against the LOO Ground Truth. Humans showed only moderate consistency with the LOO consensus (average Spearman’s $\rho = 0.330$), whereas LLMs achieved substantially higher consistency ($\rho = 0.514$). As shown in Table~\ref{tab:loo_agreement}, LLMs outperformed humans across all categories, indicating that a single LLM provides a closer approximation to the ground truth than a single human annotator.

\begin{table}[t]
\centering
\small
\resizebox{\columnwidth}{!}{
\begin{tabular}{lcc}
\toprule
\textbf{Rhetorical Strategy} & \textbf{Human vs. LOO Human GT} & \textbf{LLM vs. LOO Human GT} \\
\midrule
Causal     & 0.357 & 0.523 \\
Empirical  & 0.308 & 0.496 \\
Moral      & 0.392 & 0.609 \\
Emotional  & 0.264 & 0.427 \\
\midrule
\textbf{Average} & 0.330 & 0.514 \\
\bottomrule
\end{tabular}
}
\caption{Agreement with consensus (Spearman Correlation) between individual LLM or individual human annotators and Leave-One-Out (LOO) Human Ground Truth. LLM annotators consistently achieve higher alignment with human consensus than independent human annotators.}
\label{tab:loo_agreement}
\end{table}

Third, LLMs demonstrate greater internal consistency than human annotators. Human annotators varied widely in pairwise agreement, reflecting relatively inconsistent application of the guidelines even after the intensive training we administered. In contrast, independent LLM annotators produced more stable and coherent agreement with one another across classification schemes of varying granularity (see Table~\ref{tab:internal_consistency} in Appendix~\ref{sec:irr}). This stability suggests that, under our rhetorical strategy typology, LLM annotation is more reproducible and scalable than crowd-sourced human annotation.

\subsection{Fine-tuned Model Performance}
\label{sec:classifier}

We fine-tuned two pre-trained transformer models, RoBERTa-base \citep{liu2019robertarobustlyoptimizedbert} and LLaMA-3.2-3B-Instruct \citep{meta_llama3.2}, on individual arguments from GPT-generated debates, using LLM-based strategy labels scaled from 0 to 1 in a regression setting. RoBERTa was fine-tuned on an NVIDIA A100 GPU with a learning rate of $2 \times 10^{-5}$ and a batch size of 32. LLaMA was fine-tuned using 4-bit quantization and LoRA adapters (rank = 256, $\alpha$ = 512), with a learning rate of $4 \times 10^{-5}$, also on an A100 GPU. For all models reported in this paper, we use this same set of parameters. We evaluated the performance and topic-generalizability of the fine-tuned models in two experiments using the LLM-labeled, AI-generated dialogues. 

\vspace{-1ex}
\subsubsection{Transfer Learning Experiment on AI-generated Debates}
\label{subsec:performance}

We first evaluated how well the models generalize across topics with varying levels of exposure using the arguments from the generated debates, with each topic classified as either political or non-political. Models were trained on all arguments (N=7930) from a randomly selected 101 out of 125 political topics identified in \Cref{sec:stance_gen}, using an 8/1/1 split for training, validation, and in-domain testing. We then evaluated performance on two held-out sets: (1) all arguments (N=1528) from 20 remaining political topics for out-of-distribution (OOD) testing, and (2) all arguments (N=1962) from 25 non-political topics to assess cross-domain transfer.

We fine-tuned the model independently with three random seeds on the same training set, and report the testing set performance for each of the trained models. Each test yielded two performance scores: the Spearman rank correlation between the model's predicted rhetorical strategy score and the LLM-based scores, and the RMSE for the prediction. The scores were nearly identical across the three tests, and we report the mean correlation and mean RMSE in \cref{tab:transfer_learning}. The table reports two key findings. First, the RoBERTa model demonstrated strong predictive alignment with LLM-based scores, with exceptionally high Spearman correlations, ranging from 0.850 (cross-domain causal strategy) to 0.939 (in-domain moral strategy), and low RMSE values, ranging from 0.072 (in-domain emotional strategy) to 0.132 (cross-domain moral strategy). Second, the RoBERTa model exhibited robust transfer learning performance, with nearly identical correlations for in-domain and cross-domain evaluations, all below the 0.024 observed for the moral strategy. In sum, the results show that the two fine-tuned models are able to identify rhetorical strategies across different topics in LLM-simulated human debates. 

\vspace{-1ex}
\subsubsection{Final Model Performance with Human Validation}
\label{sec:performance}

Table 1 also shows that LLaMA under-performed RoBERTa-base, which we chose for fine-tuning on the full set of LLM-generated debate data using an 8/1/1 train/validation/test split. The model’s test performance is reported in \Cref{tab:gpt_annotation_results}. Spearman rank correlations between the model’s predictions and LLM-based scores range from 0.888 to 0.950, indicating strong alignment with the synthetic annotations. To further assess external validity, we also calculated Spearman correlations on a subset of the test data annotated by human raters. These correlations ranged from 0.607 to 0.729, providing additional evidence that the model generalizes well to human-labeled data.   

\begin{table}[!t]
\centering
\resizebox{0.98\columnwidth}{!}{
\begin{tabular}{l|cccc}
\hline
\textbf{Test Set Against} & \textbf{Causal} & \textbf{Empirical} & \textbf{Moral} & \textbf{Emotional} \\
\hline
GPT Label        & 0.888 & 0.921 & 0.950 & 0.890 \\
Human Annotation & 0.607 & 0.637 & 0.729 & 0.644 \\
\hline
\end{tabular}
}
\caption{\textbf{Model testing performance on persuasion strategy labels.} Spearman rank correlations on synthetic test set with GPT-annotated labels and human annotations.}
\label{tab:gpt_annotation_results}
\end{table}

On the human-annotated presidential debate dataset, our model also demonstrates strong transfer learning performance, with correlations between model scores and human labels ranging from 0.567 to 0.618 (see \Cref{tab:roberta_human_corr} in \Cref{sec:human_valid_president}).

\subsection{Validity Check with External Corpora}
\label{sec:valid}

To further evaluate external validity, we tested the performance of our classifier on external datasets containing binary human annotations for rhetorical strategies that are relevant to our typology. \Cref{tab:external_valid} reports the mean difference between our model’s scores and the dataset binary labels, with two-sample t-tests (see \Cref{sec:external_def} for details, including the definitions of the relevant labels). 

The results reveal two key patterns. First, our model performs best on debate-like arguments with formal argumentative structures, such as those in Presidential debates, compared to less structured contexts like charity appeals or rental requests. In the debate dataset, the mean score differences range from 0.1 to 0.409 across strategies. Second, the model effectively detects rhetorical patterns associated with specific persuasion strategies, independent of the substantive content. For example, strategies like \textit{slippery slope} and \textit{false cause}, though fallacious, both entail causal reasoning. The model is able to distinguish these based~on their argumentative form rather than the specific content of the argument, indicating the capacity to generalize across structurally similar persuasive techniques.

\begin{table}[H]
\centering
\resizebox{\columnwidth}{!}{
\begin{tabular}{p{1.9cm}p{8.5cm}p{3cm}P{4.2cm}}
\hline
\textbf{Strategy} & \textbf{Context (with Dataset Citation)} & \textbf{Relevant Label} & \textbf{Pos(1) v.s. Neg(0) Mean} \\
\hline
\textbf{Causal} & \multirow{2}{10.5cm}{\makecell[l]{Fallacious Argument in Presidential Debate\\ \citep{ijcai2022p575}}} & Slippery Slope & $0.409^{***}$ \\
&& False Cause & $0.193^{***}$ \\\cline{2-4}
& Charity Donation Requests \citep{wang-etal-2019-persuasion} & Logical Appeal & $0.047^{***}$ \\
\hline
\textbf{Empirical} & Charity Donation Requests \citep{wang-etal-2019-persuasion} & Credibility & $0.147^{***}$ \\
& Renting and Pizza Requests \citep{Chen_Yang_2021} & Evidence & $0.059^{***}$ \\
& Fallacious Argument in Presidential Debate \citep{ijcai2022p575} & Appeal to Authority & $0.100^{***}$ \\
\hline
\textbf{Emotional} &Fallacious Argument in Presidential Debate \citep{ijcai2022p575} & Appeal to Emotion & $0.200^{***}$ \\
& Charity Donation Requests \citep{wang-etal-2019-persuasion} & Personal Story & $0.160^{***}$ \\
\hline
\textbf{Moral} & Online Petitions \citep{kim-etal-2024-moral} & Moral Emotion & $0.225^{***}$ \\
\hline
\end{tabular}
}
\caption{\textbf{External Validity Test of the Strategy Models.} The table reports the average difference in model-predicted persuasion scores between positively labeled and other examples across external datasets.}
\label{tab:external_valid}
\end{table}

\vspace{-2ex}
\section{Case Studies of Two Applications}

Our classifier’s usefulness is demonstrated in two applications: 1) improving the performance of a model for predicting the persuasiveness of an argument, and 2) measuring temporal changes in rhetorical strategies in partisan political discourse.

\begin{table*}[!t]
\centering
\resizebox{0.92\textwidth}{!}{
\begin{tabular}{l|cc|cc|cc|cc|cc}
\toprule
& \multicolumn{2}{c|}{ConvArg (1038)} & \multicolumn{2}{c|}{IBM-30k (30497)} & \multicolumn{2}{c|}{IBM-5.3k (5298)} & \multicolumn{2}{c|}{IAC (4939)} & \multicolumn{2}{c}{IDEA (1205)} \\
\midrule
\midrule
&  \textbf{Spearman's}$\mathbf{\rho}$ $\uparrow$& \textbf{RMSE} $\downarrow$& \textbf{Spearman's} $\mathbf{\rho}$ $\uparrow$& \textbf{RMSE} $\downarrow$ &  \textbf{Spearman's} $\mathbf{\rho}$ $\uparrow$& \textbf{RMSE} $\downarrow$ & \textbf{Spearman's} $\mathbf{\rho}$ $\uparrow$& \textbf{RMSE} $\downarrow$ &  \textbf{Spearman's} $\mathbf{\rho}$ $\uparrow$& \textbf{RMSE}  $\downarrow$\\
\midrule
\textbf{Within Dataset - Vanilla}& 0.647 (0.012)& 0.265 (0.004)&0.502 (0.004)& 0.176 (0.004)& 0.456 (0.010)&0.204 (0.004)& 0.670 (0.000)& 0.188 (0.008)&0.263 (0.021)&0.280 (0.007)\\
\textbf{Within Dataset-Strategy}& \textbf{0.680 }(0.009)&\textbf{0.255} (0.003)&\textbf{0.516}(0.005) & \textbf{0.167} (0.003)&\textbf{0.478} (0.009)& \textbf{0.188} (0.004)&\textbf{0.678} (0.003)& \textbf{0.171}  (0.005)&\textbf{0.337 }(0.036)&\textbf{0.264} (0.007) \\
\midrule
\midrule
\textbf{Cross Dataset - Vanilla}&0.300 (0.018)& 0.335 (0.003)& 0.290 (0.005)&0.247 (0.019)&0.380 (0.005)& 0.345 (0.004&0.349 (0.003)&0.283 (0.004)&  0.052 (0.010)& 0.396 (0.005) \\
\textbf{Cross Dataset-Strategy}& \textbf{0.341} (0.016)&\textbf{0.326} (0.001)& \textbf{0.309} (0.009)&\textbf{0.218} (0.012)&\textbf{0.400} (0.005)&\textbf{0.335} (0.004)& \textbf{0.389} (0.014)&\textbf{0.257 }(0.008)& \textbf{0.053 } (0.009)& \textbf{0.395 }(0.004)\\
\bottomrule
\end{tabular}
}
\caption{\textbf{Persuasiveness Score Performance.} Performance of models with and without the incorporation of rhetorical strategies, evaluated within and across datasets (higher $\rho$, lower RMSE are better). "Vanilla" refers to the condition without incorporation of labels for rhetorical strategies. Results are averaged over three fine-tuning runs (mean ± SD). Full results with performance differences and standard errors are reported in the Appendix.}
\label{tab:persuassiveness}
\end{table*}

\vspace{-1.5ex}

\subsection{Persuasiveness Score Prediction }
Changing someone’s opinion is a common goal in contexts ranging from political and marketing campaigns to everyday interactions. This has made the study of what makes an argument persuasive a longstanding area of interest \citep{persuasionbook,habernal_second,wang-etal-2019-persuasion,winning,toledo2019automatic}. 
We illustrate the usefulness of the classifier model by testing whether knowledge of an argument's rhetorical strategy can improve performance in predicting the persuasiveness of the argument. To test this, we conducted experiments across five datasets drawn from diverse topical domains, providing a broad testbed for evaluating both domain-specific and cross-domain performance. Each dataset contains arguments whose persuasiveness was assessed by human judges. The size of each dataset is shown in Table \ref{tab:persuassiveness}. A detailed description of the datasets and the evaluation of persuasiveness is provided in Appendix~\ref{sec:datapers}.

We tested model performance in two settings: within and across topical domains, corresponding to five datasets with qualitatively different argumentation. The within domain analysis assesses performance in domain-specific contexts using an 8/1/1 train/validation/test split for each domain.  We also tested the model’s ability to generalize across domains with differing linguistic features. In the cross-domain setting, we fine-tuned the model on four of the five datasets and tested on the held-out fifth dataset. For each argument in each dataset, we applied the RoBERTa-based classifier trained on GPT-generated debate data to predict the four strategy scores. For both tasks, we used mean squared error to fine-tune a BERT-base-uncased model and project the resulting representation to a 128-dimensional vector. We then projected the four strategy scores into a 32-dimensional vector, concatenated this with the textual representation, and passed the combined vector through a 64-dimensional projection layer to score the persuasiveness of the argument. We measured performance using two complementary metrics, Spearman correlation between predicted and ground-truth persuasion scores and RMSE. We then compared performance between two conditions, with and without inclusion of predicted strategy scores.


\Cref{tab:persuassiveness} reports small but consistent improvements in predicting persuasiveness when incorporating rhetorical strategy. Within-domain, the strategy features increased the correlation with ground-truth persuasiveness scores by a relative $8.40\%$ (absolute $0.03$), with a relative $6.30\%$ (absolute $0.014$) decrease in RMSE, indicating better alignment with human judgments. In the more challenging cross-domain setting, we observe a relative $7.77\%$ (absolute $0.024$) increase in correlation and a relative $6.16\%$ (absolute $0.015$) reduction in RMSE. These improvements suggest that the strategy features not only improve prediction within a given domain but also in topical contexts other than those on which the model was trained. This case study highlights the value of using rhetorical strategies for more robust, generalizable analysis of persuasive arguments.

\vspace{-1.5ex}
\subsection{U.S. Presidential Debates as an indicator of Affective Polarization}
\label{subsec:presidential}

The postwar increase in affective partisan polarization in the U. S. is evident not only in the voting population but also among political elites \citep{Enders_2021}. This suggests the hypothesis that political elites have shifted from cognitive discourse in the relatively bipartisan Eisenhower years to increasingly affective discourse today. Our model offers the opportunity to test this hypothesis by analyzing the transcripts of U. S. Presidential debates, going back to the inaugural Kennedy-Nixon debate in 1960 \citep{Martherus_2020}. The hypothesis could also be tested using the \textit{Congressional Record}, campaign ads, and stump speeches, but Presidential debates afford unique access to elite argumentation that targets a national audience, is focused exclusively on politically salient controversies, and follows institutional procedures that have remained relatively constant over time. 

\begin{figure}[t]
    \centering
    \begin{subfigure}[t]{0.48\columnwidth}
        \centering
        \includegraphics[width=\linewidth]{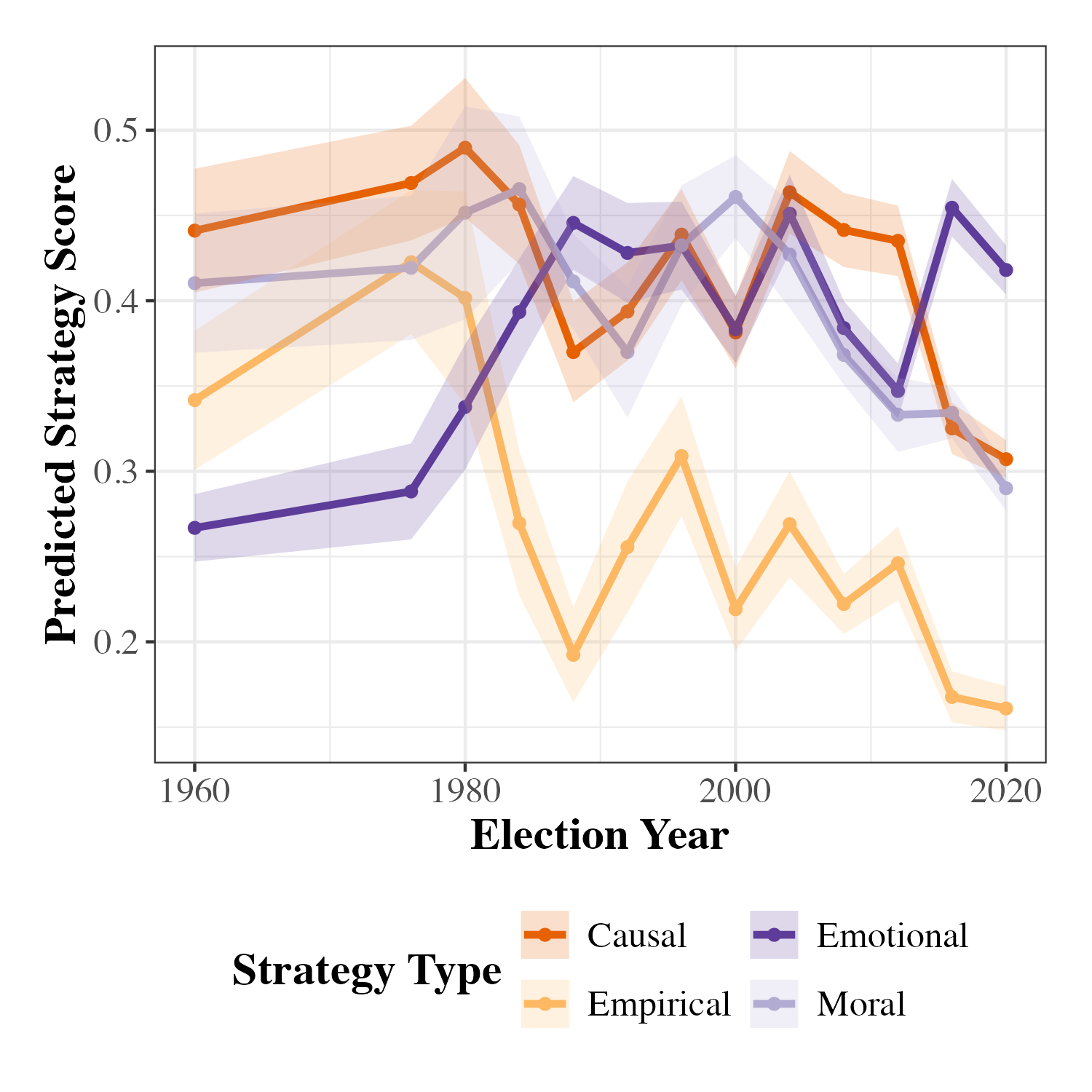}
    \end{subfigure}
    \hfill
    \begin{subfigure}[t]{0.48\columnwidth}
        \centering
        \includegraphics[width=\linewidth]{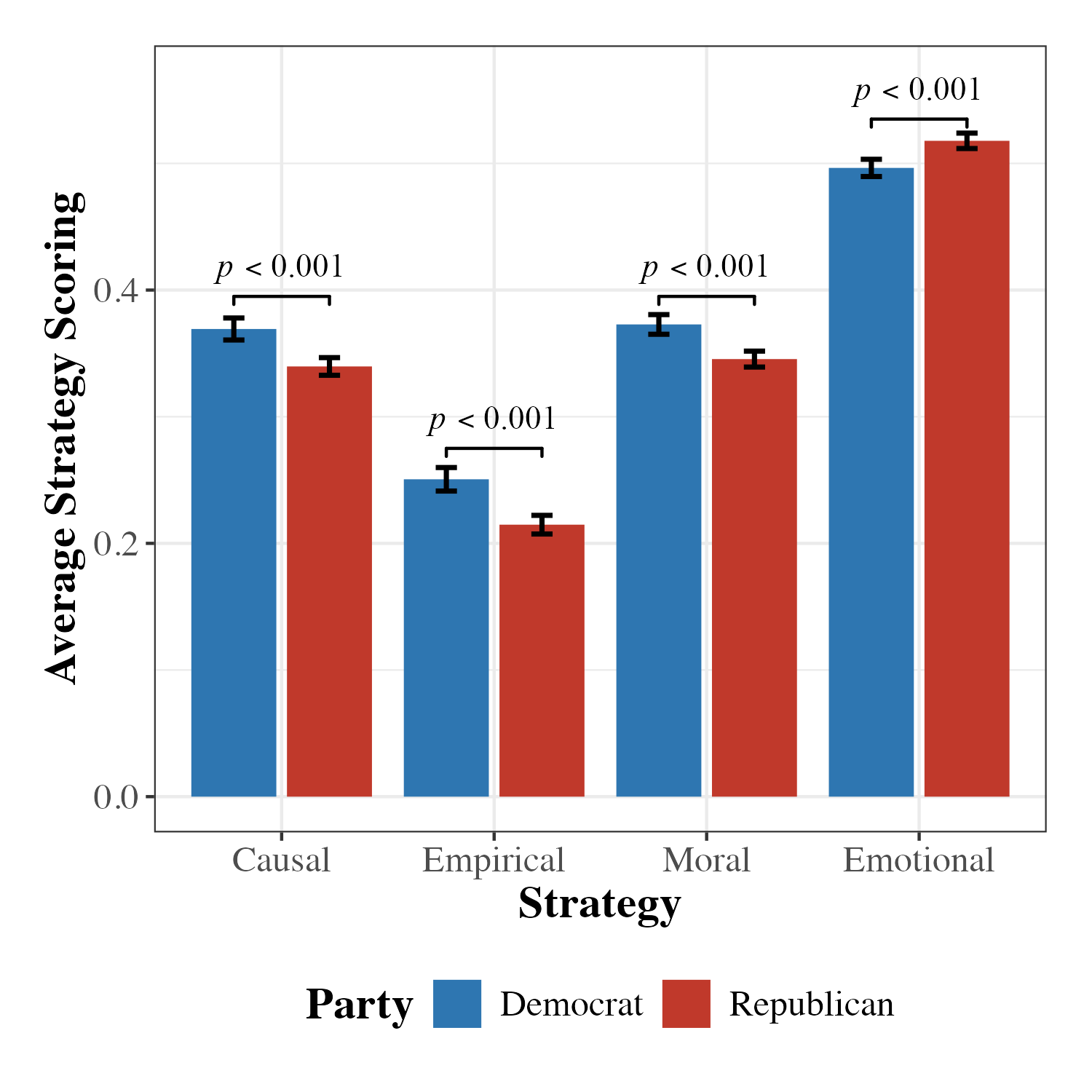}
    \end{subfigure}
    \caption{\textbf{Rhetorical strategies in U.S. Presidential debates.} Left: temporal trends (1960–2020). Right: partisan differences.}
    \label{fig:debate_trends}
    \vspace{-2ex}
\end{figure}

We measured temporal trends and partisan differences in  rhetorical strategies at the argument level, defined as a continuous, uninterrupted string by a single speaker, with at least five words. For each argument, we applied our classifiers trained in \Cref{sec:performance} to predict each of the strategies. For comparability, we limited the analysis to general election candidates from the two major political parties and excluded Vice Presidential and primary debates, which differ in format and are only available for certain years. The debate corpus for analysis covers 13 U.S. presidential elections since 1960 (no debates were held in 1968 and 1972), totaling 3,307 arguments.

\subsubsection{Temporal trends}

\Cref{fig:debate_trends} (left) reports predicted strategy scores across Presidential debates by election year, with 95\% confidence intervals. Empirical strategies (gold) show a consistent decline while emotional appeals (purple) increased, suggesting a shift from evidence-based cognitive arguments to affective rhetorical strategies. This trend is confirmed by a linear model using a single aggregated measure of cognitive (the mean of causal and empirical scores) minus affective (the mean of emotional and moral scores) on each argument. Affective scores increased relative to cognitive by 0.0025 per year ($p < 0.001$), approximately a 0.01 increase per four-year election cycle beginning in 1976. 

\vspace{-1ex}
\subsubsection{Partisan differences}

\Cref{fig:debate_trends} (right) reports temporally aggregated partisan differences in rhetorical strategies in Presidential de-bates. Compared to Democrats, Republican candidates relied more on emotional strategies ($\Delta$ = 0.021, $p<0.001$), and less on causal ($\Delta$ = 0.029, $p<0.001$), empirical ($\Delta$ = 0.036, $p<0.001$), and moral strategies ($\Delta$ = 0.027, $p<0.001$). However, across all elections since 1960, both Democrats ($\Delta$ = 0.246, \textit{p} < 0.001) and Republicans ($\Delta$ = 0.303, \textit{p} < 0.001) relied far more on emotional than on empirical arguments ($\Delta$ = 0.277, \textit{p} < 0.001). For election-specific results, see  \Cref{sec:debate_individual}.


\vspace{-1ex}
\section{Conclusion}

Large-scale identification of rhetorical strategies has been hindered by the limitations of human annotation, including high cost, inconsistency, and limited scalability due to cognitive demands. To address this, we used a novel framework that leverages large language models to generate and annotate four rhetorical strategies in debate data. These synthetic labels are validated by simulated~LLM personas and human annotators, enabling the fine-tuning of a robust rhetorical classifier that generalizes across topical domains. We demonstrate~its utility in two applications: improving~persuasiveness prediction and revealing the rise in affective appeals and decline in empirical arguments in U.S. Presidential debates from 1960 to 2020.

\section*{Limitations and Future Work}

We note several limitations of our current study. First, we generated and evaluated data in English, which may overlook persuasion strategies that manifest differently across other languages and cultures. Future research is needed to extend our framework to multiple languages and cross-cultural comparisons, such as between the East and West and between individualist vs. collectivist societies. Second, our training data simulates only debate settings, but we can potentially improve transferability by incorporating simulations from other persuasive contexts such as advertising or fundraising. Third, we used four rhetorical strategies that were more refined than previous typologies, but future research is needed to test more fine-grained distinctions corresponding to specific emotions (e.g. indignation) or types of evidence (e.g. eye-witness or statistical). The modest improvement we observed in persuasiveness prediction may be amplified by discovering specific strategies that are uniquely effective in certain contexts. 

Another limitation is the focus on persuasion, but rhetorical strategies may also influence information diffusion. Future research is needed to identify strategies that trigger virality on social media. We also did not take veracity into account. Going forward, a promising direction is to compare rhetorical strategies used in arguments that are truthful, intentionally misleading, or misinformed. For example, are affective strategies key to the manipulation and dissemination of falsehoods, with potential applications to mass persuasion processes and the spread of disinformation.

\section*{Potential Risks and Ethical Considerations}

The synthetic debate dialogues generated and analyzed in this study were developed solely for research and model training purposes. While our framework offers scalability, flexibility, and high accuracy for rhetorical strategy analysis, we acknowledge the potential for misuse. As with many advances in natural language processing, similar frameworks could be repurposed by malicious actors to generate or evaluate manipulative and misleading content. However, this risk is not unique to our study and reflects broader concerns about the dangers and misuse of generative AI technologies.

Human annotation studies in this project were reviewed and approved by Cornell University’s Institutional Review Board (IRB), which granted an exemption under Protocol Number IRB0149357. Annotators for rhetorical strategies were recruited through the Prolific platform, participated with informed consent, and were compensated in line with the platform’s pay guidelines. No personally identifiable information was collected during the human annotation study. Participants were only associated with platform-assigned anonymous IDs used solely for payment purposes.

All external datasets used for model evaluation are publicly available to the research community. To promote transparency and facilitate future research, we will publicly release the full synthetic dataset and associated model outputs upon publication.

\section*{Acknowledgements}
This work is supported in part by NSF Awards 2242073 and 2242072, by the U.S. National Library of Medicine (R01LM013833), and by a grant from the John Templeton Foundation.

\bibliography{main}
\appendix

\clearpage

\onecolumn

\section{Persuasion Strategies Literature}
\label{sec:strategytable}

\begin{table*}[!h]
\centering
\resizebox{0.95\textwidth}{!}{%
\begin{tabular}{p{0.15\linewidth}p{0.20\linewidth}p{0.45\linewidth}p{0.2\linewidth}}
\hline
\textbf{Our Typology} & \textbf{Related Concept} & \textbf{Related Definition} & \textbf{Source} \\ \hline

\multirow{7}{*}{Causal} 
& \makecell[l]{Reason} & Provides a justification for an argumentative point based on additional argumentation schemes, e.g., causal reasoning or argument absurdity. & \makecell[l]{\citet{anand_2011} \\ \citet{iyer_2019}} \\ \cline{2-4}
& Reframing & Reframe issues through usage of analogy or metaphor & \citet{duerr2021persuasivenaturallanguagegeneration} \\ \cline{2-4}
& Counter-arguments & Predict possible opposing opinions and prepare rebuttal arguments. Increase persuasiveness by addressing the audience’s doubts and concerns. & \citet{jin-etal-2024-persuading} \\ \cline{2-4}
& Pro and Con & Provide the audience with an analysis of the pros and cons of the point of view, letting them understand why your point of view is more advantageous for them & \citet{jin-etal-2024-persuading} \\ \cline{2-4}
& \multirow{2}{*}{Logos} & Appeals to logical reason & \citet{cabrio_proceedings_2018} \\ \cline{3-4}
& & Appeal to the rationality of the audience through logical reasoning & \citet{hidey-etal-2017-analyzing} \\ \hline

\multirow{4}{*}{Empirical} 
& Evidence & Using supporting evidence such as statistics, examples, facts & \citet{shaikh-etal-2020-examining} \\ \cline{2-4}
& Concreteness & The use of facts or evidence & \citet{yang-etal-2019-lets} \\ \cline{2-4}
& \multirow{2}{*}{Logos} & Appealing to the audience through reasoning or logic, by citing facts and statistics, historical and literal analogies. & \citet{marro-etal-2022-graph} \\ \cline{3-4}
& & Factual argumentation & \citet{abbott-etal-2016-internet} \\ \hline

\multirow{5}{*}{Emotional} 
& Empathy & Encourage the audience to connect with someone else’s emotional state & \citet{anand_2011} \\ \cline{2-4}
& \multirow{2}{*}{Pathos} & Persuade an audience by appealing to their emotions & \citet{marro-etal-2022-graph} \\ \cline{3-4}
& & Aims at putting the audience in a certain frame of mind, appealing to emotions, or more generally touching upon topics in which the audience can somehow identify & \citet{hidey-etal-2017-analyzing} \\ \cline{2-4}
& \multirow{2}{*}{Emotion} & Have recipient feel certain emotions (guilt, anger, shame, fear, pity, feeling important, content, etc.) & \citet{miceli2006emotional} \\ \cline{3-4}
& & Messages with high emotional valence and arousal & \citet{yang-etal-2019-lets} \\ \hline

\multirow{3}{*}{Moral} 
& Deontic Appeals & Mentions duties or obligations & \citet{anand_2011} \\ \cline{2-4}
& Moral Appeals & Mentions moral goodness/badness & \citet{anand_2011} \\ \cline{2-4}
& Emotion & Have recipient feel certain emotions (guilt, anger, shame, fear, pity, feeling important, content, etc.) & \citet{miceli2006emotional} \\ \hline

\end{tabular}
} 
\caption{Overview of persuasive strategies with definitions and corresponding related work. This typology connects our four-category framework with established concepts and definitions from the literature.}
\label{tab:strategies}
\end{table*}

\clearpage

\twocolumn
\balance

\section{Opposing Stances Generation}
\label{sec:topicprompt}
\subsection{Prompt for Opposing Stance Generation}
\begin{figure}[H]
    \centering
    \includegraphics[width=1\columnwidth]{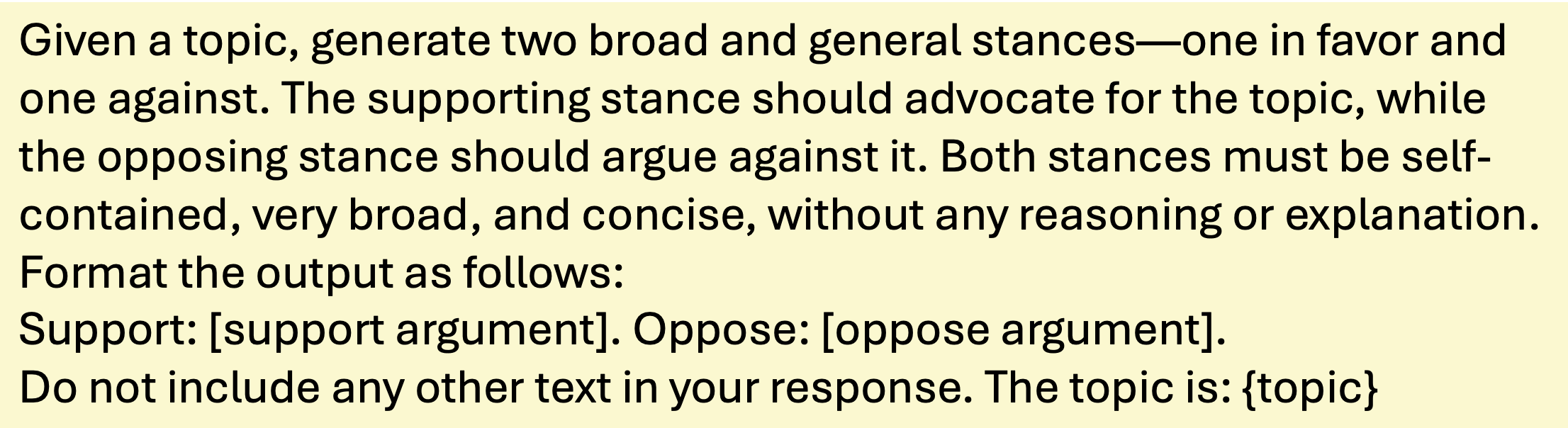}
    \caption{Prompt for opposing stance generation}
    \label{fig:topic_generation}
\end{figure}

Figure \ref{fig:topic_generation} shows the prompt used for generating opposing stances. Given a topic, we ask GPT-4o to generate broad stances both in support of and in opposition to the topic.

\vspace{-2.5ex}

\subsection{Examples of Topics and Their Generated Stances}

Table \ref{tab:topics_stances} illustrates five examples of controversial political topics, each accompanied by two opposing stances generated by GPT-4o. These stances were then used in our debate generation framework to produce persuasive arguments.

\vspace{-1.5ex}
\begin{table}[H]
    \centering
    \resizebox{\columnwidth}{!}{
    \begin{tabular}{l|ll}
        \hline
        \textbf{Topic} & \textbf{Stance 1} & \textbf{Stance 2} \\ 
        \hline
        Abortion: Late-Term &We should allow late-term abortion. & We should prohibit late-term abortion.\\ 
        \hline
        Marijuana &We should legalize marijuana. & We should not legalize marijuana.\\ 
        \hline
        Race Relations & \makecell[l]{We should prioritize improving race relations \\ to create a more inclusive society.} &  
        \makecell[l]{We should not prioritize race relations \\ above other pressing societal issues.} \\ 
        \hline
       Voter Registration & We should make voter registration automatic. & We should not make voter registration automatic.\\ 
        \hline
        Universal Health Care & We should implement universal health care.
 &  We should not implement universal health care.\\ 
        \hline
    \end{tabular}}
    \caption{Five examples of controversial topics and their generated opposing stances.}
    \label{tab:topics_stances}
\end{table}

\section{Prompts to Generate Debates}
\label{sec:prompts_debate}

\subsection{Utterance Generation Prompts}

\begin{figure}[H]
    \centering
    \begin{adjustbox}{width=1\linewidth}
    \includegraphics{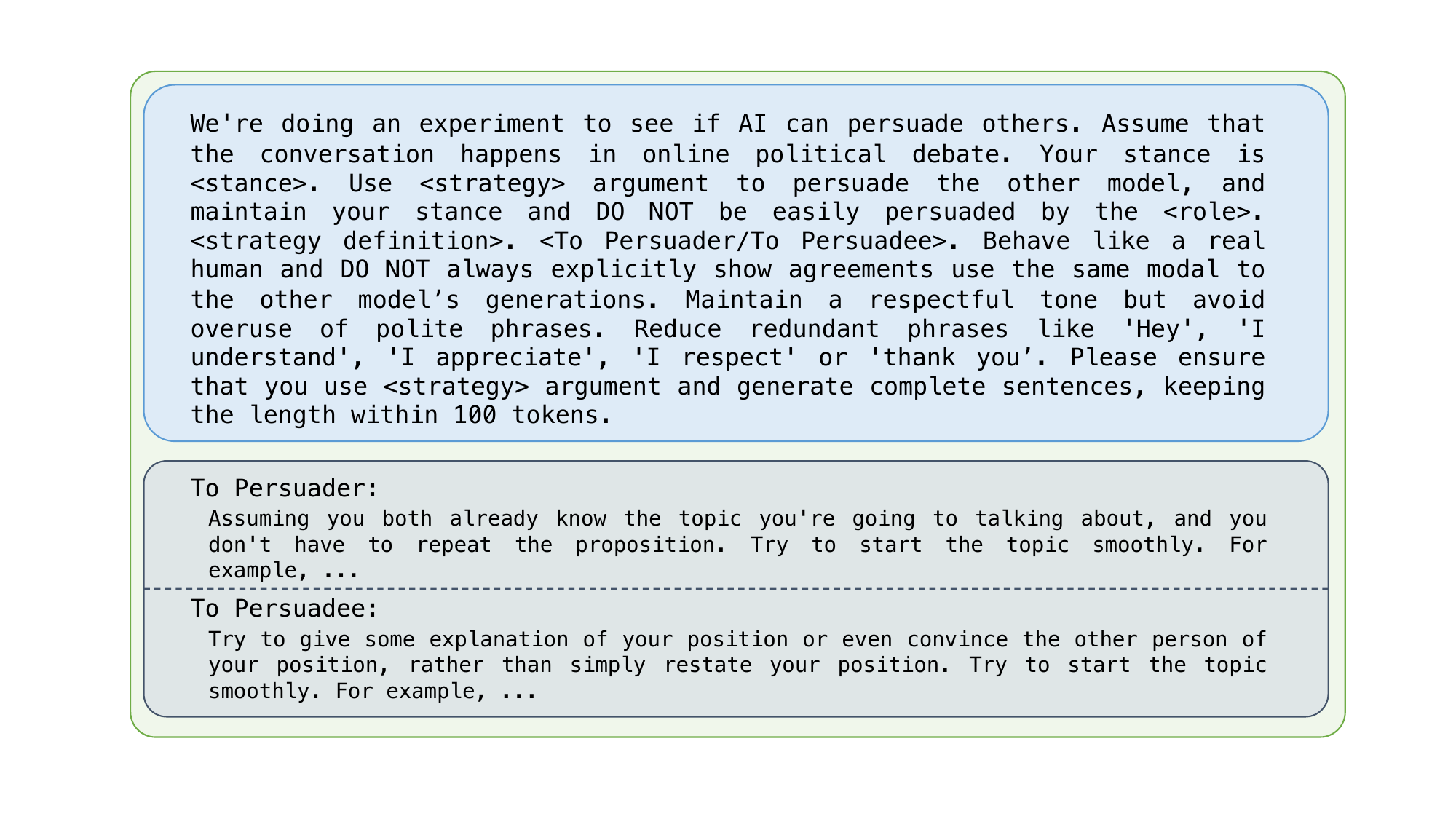}
    \end{adjustbox}
    \caption{Prompt for generating debate utterance with a certain strategy.}
    \label{fig:positive_gen}
\end{figure}

\begin{figure}[H]
    \centering
    \includegraphics[width=1\linewidth]{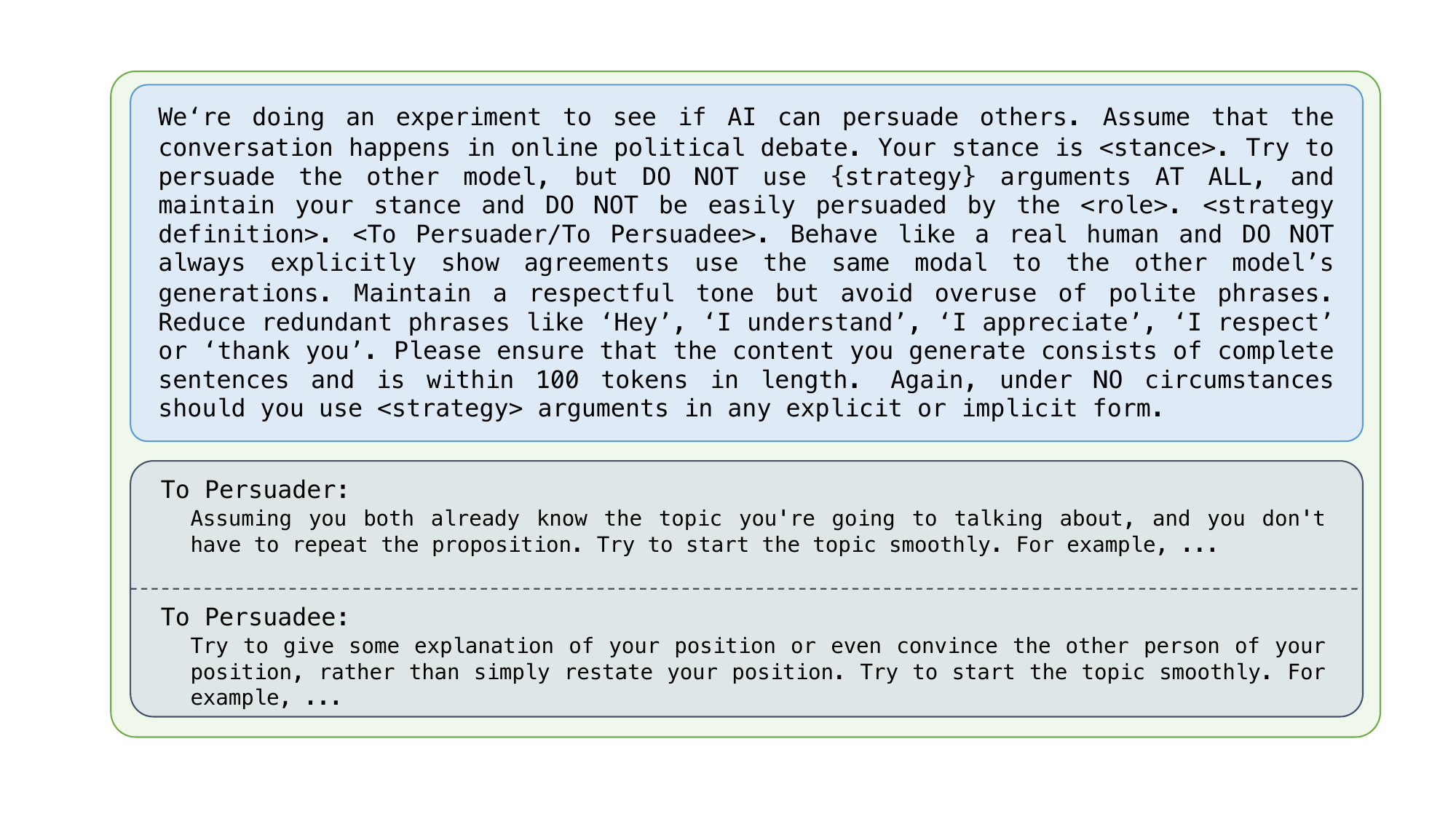}
    \caption{Prompt for generating debate utterance without a certain strategy.}
    \label{fig:negative_gen}
\end{figure}

We employed distinct prompts to instruct the agents either to adopt a specific persuasion strategy or to avoid it. The prompts used for generating debate utterances under these two conditions are illustrated in \cref{fig:positive_gen} and \cref{fig:negative_gen}, respectively. For simplicity, only the modified sections of the prompt—highlighted in blue—are shown; these reflect changes made to the original framework proposed by \citet{Ma_et_al_2025}. The unmodified portions (shown in grey), which follow the original prompt structure, are omitted for brevity.

\vspace{-2.5ex}

\begin{table*}[!t]
\centering
    \resizebox{0.9\textwidth}{!}{
\begin{tabular}{>{\raggedright\arraybackslash}p{2.5cm} >{\raggedright\arraybackslash}p{18cm}}  
\toprule
\textbf{Strategy} & \textbf{Examples} \\
\midrule
\multirow{2}{*}{Causal}   & 
Allowing prisoners to choose death reduces public pressure to improve the prison system. \\
\cmidrule(lr){2-2}
& Mandatory vaccination could result in rich countries hoarding vaccines for their population. This could make vaccines inaccessible or unaffordable for poorer countries. \\
\midrule
\multirow{2}{*}{Empirical} & 
The issue of animal extinction could be largely fixed with lab-grown meat. US consultancy firm Kearney suggests that 35\% of all meat consumed globally will be cell-based by 2040. \\
\cmidrule(lr){2-2}
& Research has estimated that many death row inmates were wrongly convicted and could have been exonerated. \\
\midrule
\multirow{2}{*}{Moral}  & 
It is a duty of the state to protect its citizens from life-threatening diseases such as COVID-19. \\
\cmidrule(lr){2-2}
& It's unfair that families of prisoners can't see prisoners; it's also unfair how they're more at risk from COVID-19. \\
\midrule
\multirow{2}{*}{Emotional} & 
Gay marriage is a lifestyle choice. It may be considered 'unnatural', but that is between that person and his/her love interest. Love is all some people have... You can't take that one given right away because it makes you uncomfortable. They want acceptance and understanding. Let them be happy or just ignore it. You don't choose to be gay either. Who would choose to live that way? They are constantly being harassed and can't be with their loved one. It's unfortunate and cruel. Please be respectful of them. They have done nothing wrong, God created them that way. \\
\cmidrule(lr){2-2}
& These players are earning disgusting weekly salaries and the NHS is on its knees and the staff are putting their lives at risk whilst the footballers stay at home drinking Molt! \\
\bottomrule
\end{tabular}}
\caption{Two Examples of Each Strategy Used in GPT-4o and Human Annotations.}
\label{tab:strategy-examples}
\end{table*}

\subsection{Strategy Refinement Prompt}

\begin{figure}[H]
    \centering
    \includegraphics[width=1\linewidth]{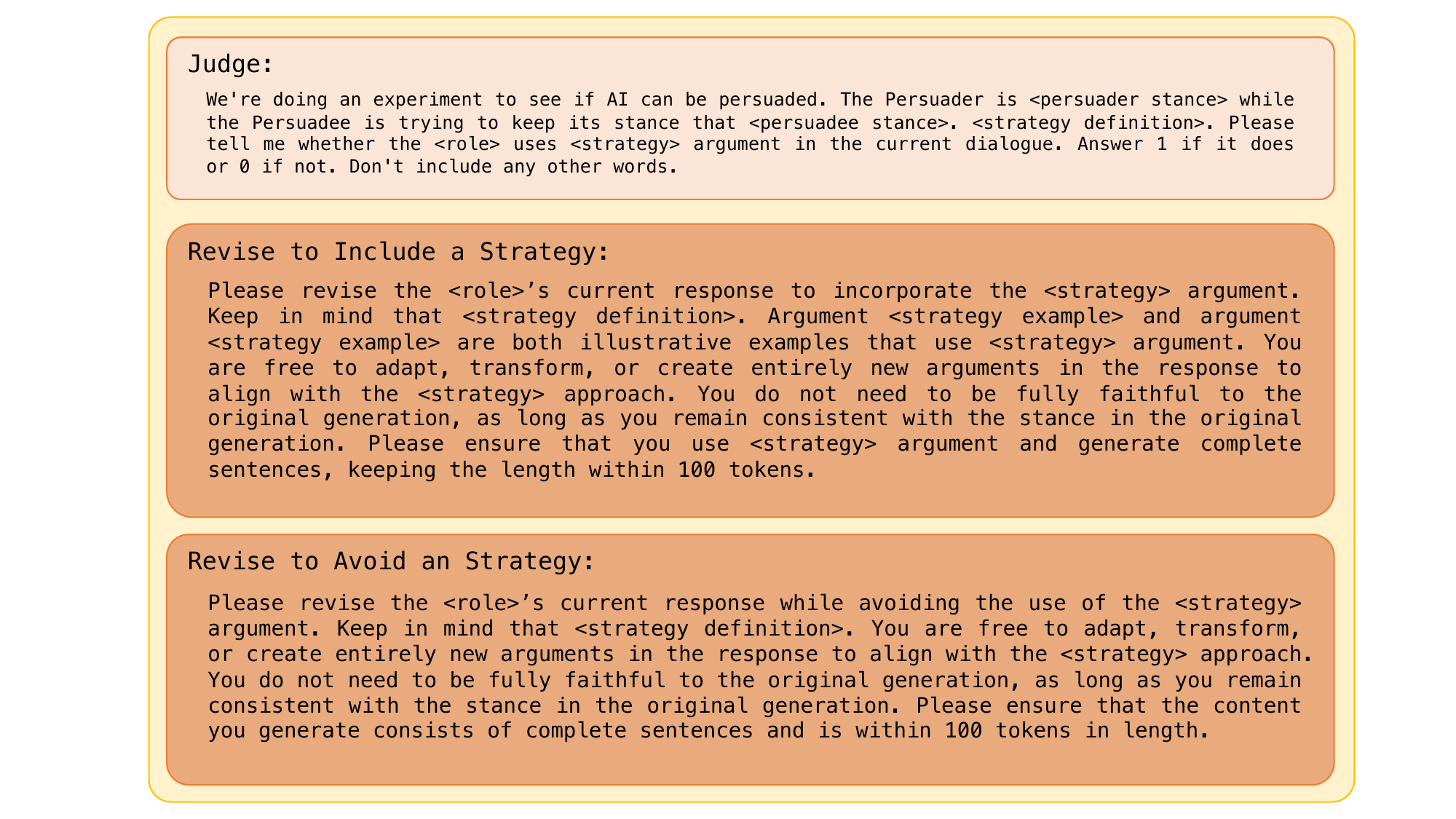}
    \caption{Prompts used in the detect-and-revise pipeline. }
    \label{fig:detect_revise}
\end{figure}

There are two steps involved in the strategy refinement. The first prompt evaluates whether the generated utterance follows the assigned persuasion strategy, and the second prompt instructs the model to revise the utterance if it fails to meet the strategy condition. The whole detect-and-revise process is repeated at most 2 times for each individual utterance. The prompts are shown in \cref{fig:detect_revise}.

\begin{figure}[H]
    \centering
    \includegraphics[width=0.9\linewidth]{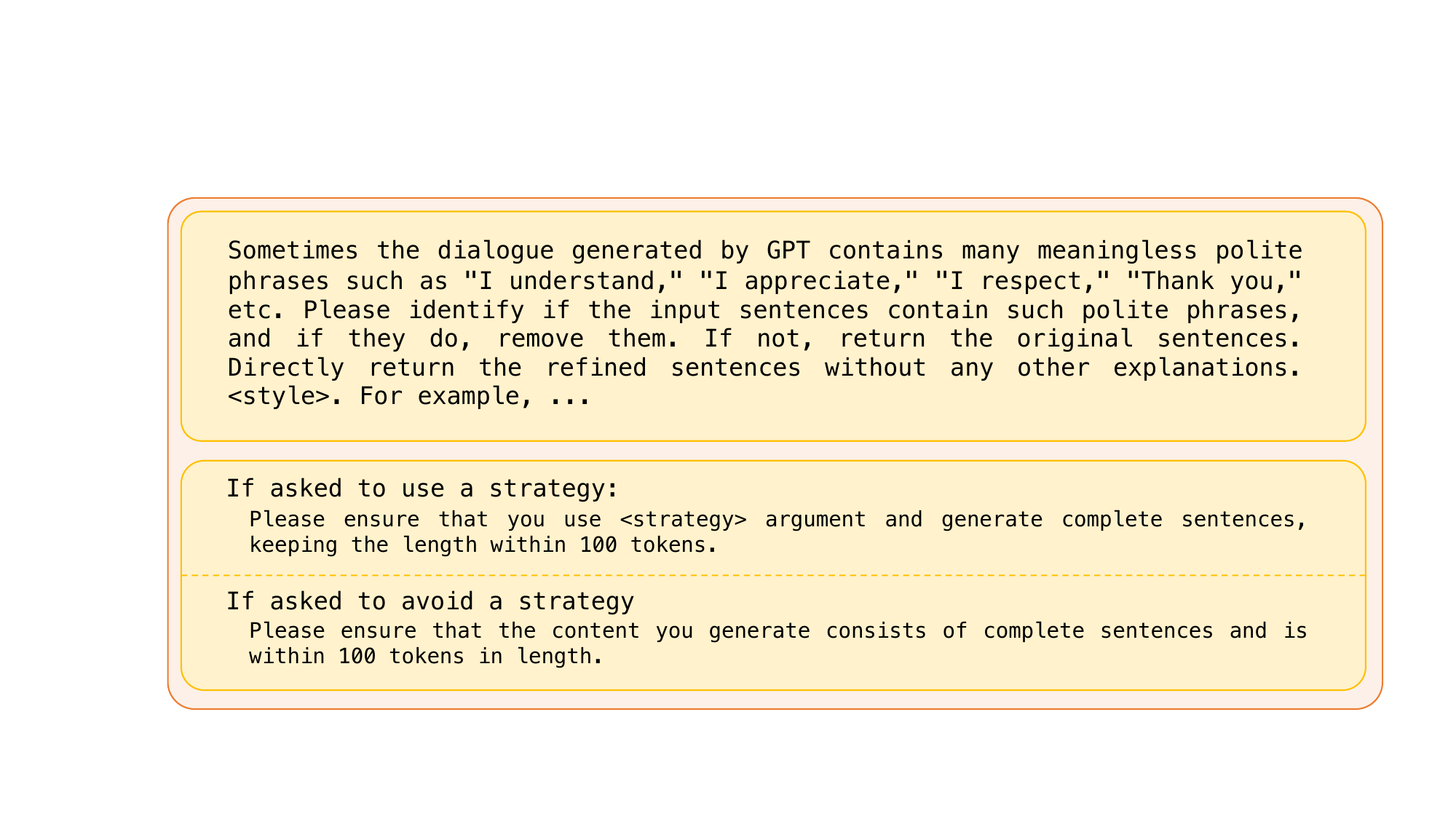}
    \caption{Prompts used to revise the individual utterance to eliminate redundancy.}
    \label{fig:refine}
\end{figure}

After the utterance has been revised for certain strategy, the model was evaluated for redundancy. And the prompt is shown in \cref{fig:refine} 

\newpage
\balance

\subsection{Round-level Refinement}

\begin{figure}[H]
    \centering
    \includegraphics[width=0.9\linewidth]{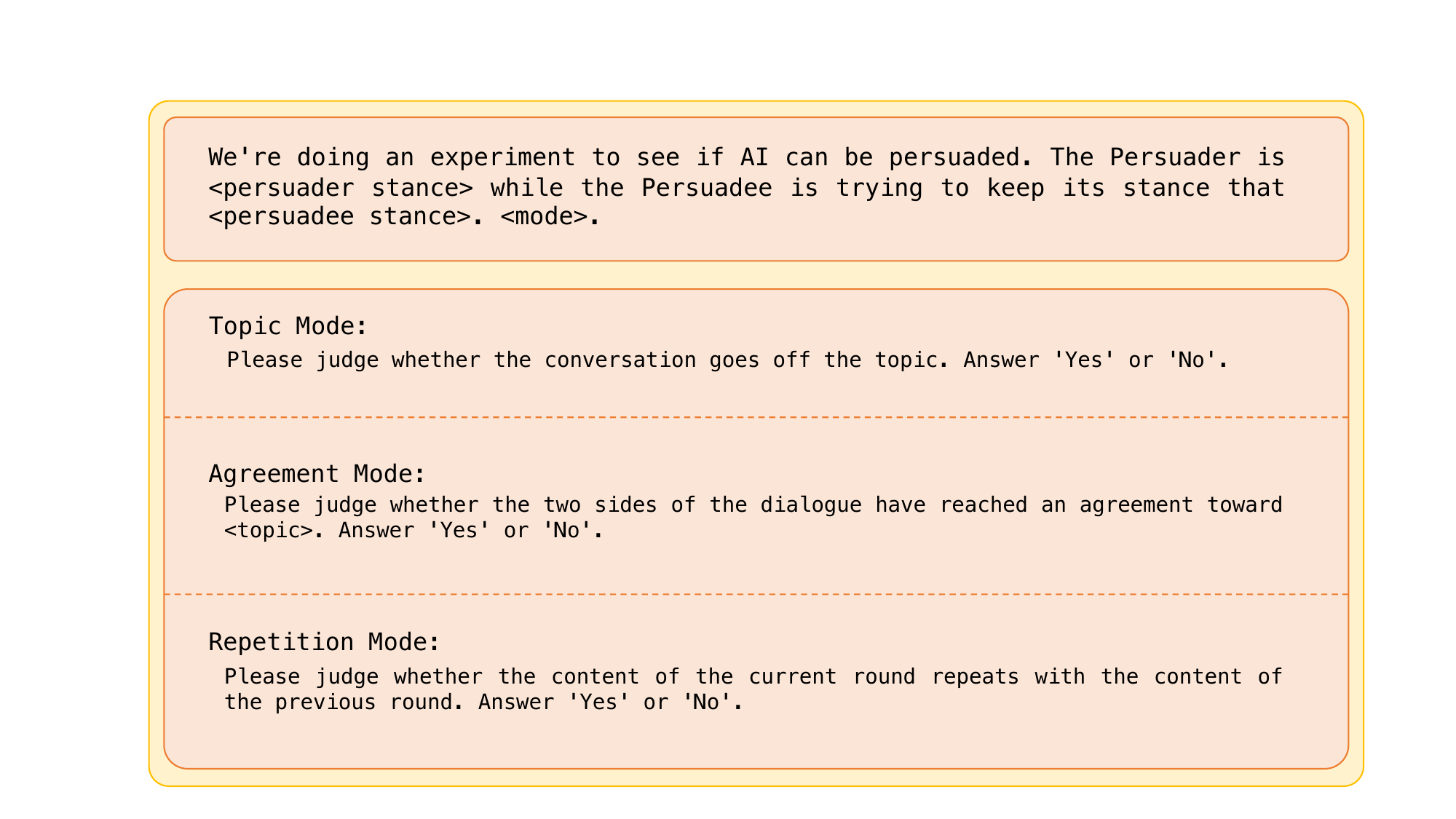}
    \caption{Prompts used to evaluate topic consistency, repetition and agent agreement for each round of arguments to improve generation quality.}
    \label{fig:global_refine}
\end{figure}

The two utterances in each debate round will be evaluated for topic consistency and repetition. If the round of utterances goes off topic or demonstrate strong repetition with the previous round, we instruct the model to re-do the generation for this round. In addition, we detect whether the two agents have reached consensus, and we stop the generation once a consensus is reached. Prompts to judge on these factors are shown in \cref{fig:global_refine}

\section{Annotation Prompt}

\begin{figure}[H]
    \centering
\includegraphics[width=0.6\linewidth]{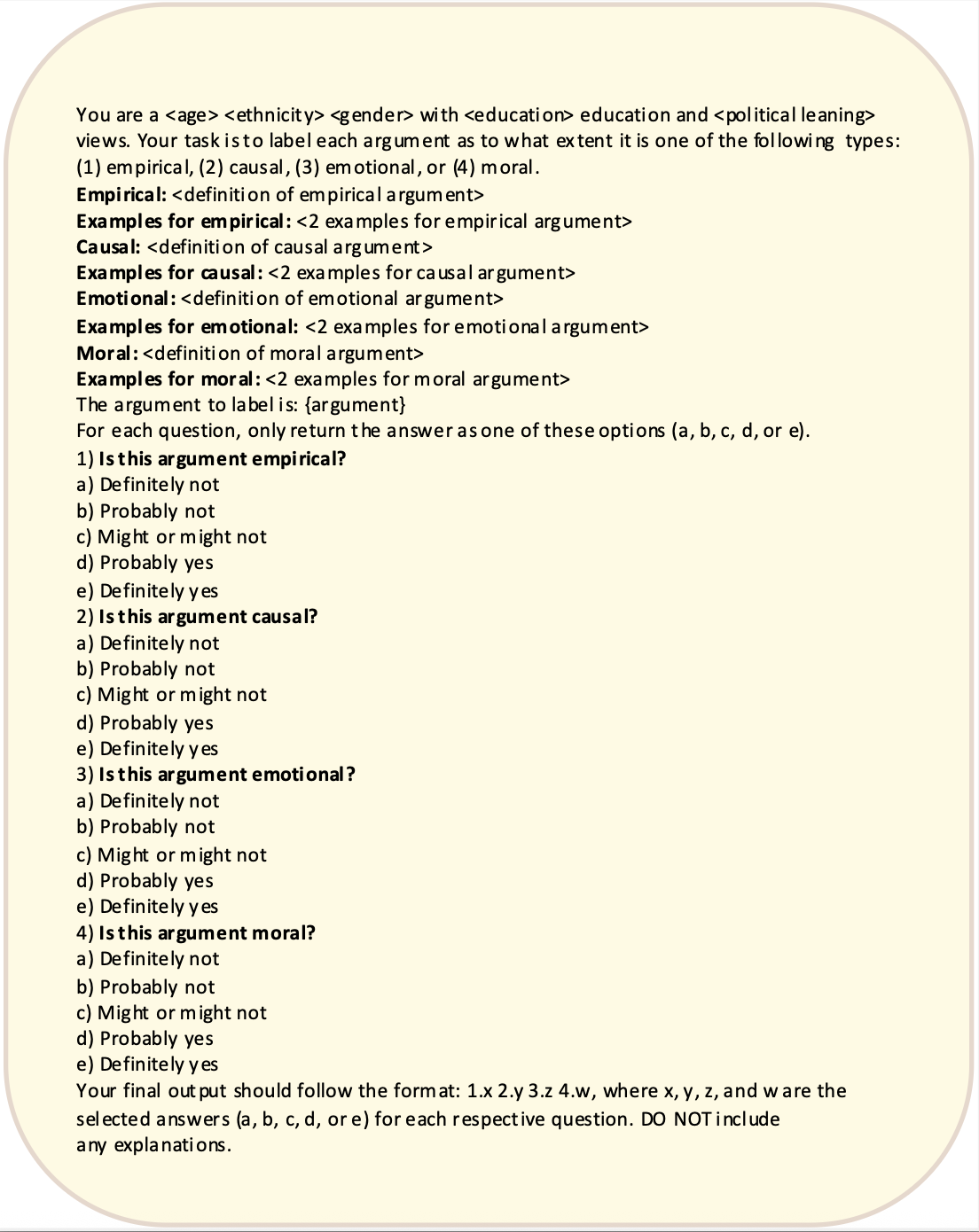}
    \caption{Prompt used for strategy annotation by GPT-4o}
    \label{fig:strategy-annotation}
\end{figure}

Figure \ref{fig:strategy-annotation}  presents the prompt used for strategy labeling by GPT-4o agents adopting different personas. The definitions of each strategy correspond to those provided in section \ref{sec:def}. We used two examples for each strategy annotated, which are provided in Table \ref{tab:strategy-examples}.
\label{sec:annptationprompt}

\section{Details of Persona Construction}
\label{sec:personaapp}

Table \ref{tab:persona_attributes} shows the persona attributes and possible values for each group, which are sampled based on U.S. Census  \citep{usbureaudata2025}, American Council on Education Statistics  \citep{acedata2024}, and Pew Research Center  \citep{pewdata2024}. To simulate a more realistic online user population, we excluded individuals under age 15 and over 89.

\begin{table}[!h]
\centering
\resizebox{0.9\linewidth}{!}{
\begin{tabular}{>{\raggedright\arraybackslash}p{3cm} >{\raggedright\arraybackslash}p{9cm}}
\hline
\textbf{Persona Attribute} & \textbf{Possible Values} \\
\hline
Gender & Male, Female \\
\hline
Age Group & $< 5$, 5–9, 10–14, 15–19, 20–24, 25–29, 30–34, 35–39, 40–44, 45–49, 60–64, 65–69, 70–74, 75–79, 80–84, $\ge 85$ \\
\hline
Race & Black, White, Asian, AIAN (American Indian and Alaska Native), NHPI (Native Hawaiian and Pacific Islander) \\
\hline
Level of Education & Less than High School, High School Graduate, Some College but No Degree, Associate Degree, Bachelor's Degree, Master's Degree, Professional Degree, Doctoral Degree \\
\hline
Political Leaning & Democrat, Republican, and Independent \\
\hline
\end{tabular}
}
\caption{Persona attribute categories and value sets used for LLM simulation}
\label{tab:persona_attributes}
\end{table}

\clearpage
\onecolumn
\section{Examples of Generated Arguments}

\begin{table}[H]
\centering
\resizebox{1\textwidth}{!}{%
\begin{tabular}{lllp{16cm}}
\hline
\textbf{Strategy} & \textbf{Topic} & \textbf{Instruction} & \textbf{Generated Utterance} \\
\hline

\centering
\renewcommand{\arraystretch}{1.2} 

\multirow{4}{*}{Causal} 
& \multirow{2}{*}{We should maintain capitalism as our economic system.} 
& Use Strategy & While some companies may prioritize profit over other considerations in a capitalist system, regulations and oversight mechanisms can effectively address these issues. Government intervention ensures that businesses operate ethically, protecting worker rights and promoting sustainable practices. Competition within capitalism can drive companies to innovate and improve their social and environmental impact, leading to adaptation and progress that benefit society and the environment in the long run. \\
\cline{3-4}
& & Avoid Strategy & However, capitalism remains a predominant economic system globally. It emphasizes individualism and personal initiative, which are heavily valued in many societies. Competition is a fundamental aspect of capitalism that many find appealing. These are deeply ingrained societal values. \\
\cline{2-4}

& \multirow{2}{*}{We should not support WikiLeaks.} 
& Use Strategy & While transparency and accountability are crucial in a functioning society, supporting WikiLeaks may not always be the best approach. By releasing classified information without proper context or consideration of potential consequences, WikiLeaks can jeopardize national security and put individuals at risk. The indiscriminate release of sensitive data can harm diplomatic relations, compromise ongoing operations, and endanger lives. It is important to promote transparency through legal and ethical means that uphold the safety and well-being of all individuals involved. \\\cline{3-4}
& & Avoid Strategy & Just because an organization is well-known doesn't mean their actions are always justified or ethical. Reputation alone should not be the sole basis for supporting an entity. There are other factors to consider beyond just fame or notoriety. \\
\hline

\multirow{4}{*}{Empirical} 
& \multirow{2}{*}{We should embrace alternative medicine alongside conventional treatments.} 
& Use Strategy & Integrating alternative medicine alongside conventional treatments can provide patients with a more comprehensive care plan. Research has shown that certain alternative therapies, such as acupuncture and herbal medicine, can effectively complement traditional medicine in managing various conditions. Would you be open to considering the benefits of embracing alternative medicine alongside conventional treatments? \\\cline{3-4}
& & Avoid Strategy & However, integrating alternative therapies doesn't necessarily mean abandoning traditional practices. Instead, it can enhance the overall efficacy of treatment by addressing aspects that conventional methods may not cover. It's like adding different colors to a painting to create a richer, more vibrant masterpiece. \\
\cline{2-4}

& \multirow{2}{*}{We should not prioritize the expansion of civil liberties.} 
& Use Strategy & While it's important to consider civil liberties, there is empirical evidence suggesting that certain restrictions can benefit the populace in crisis situations. For instance, during the COVID-19 pandemic, a study published in The Lancet found that mask mandates led to a significant decrease in infection rates. Similarly, during the 2011 London riots, temporary curfews were imposed, leading to a 70\% reduction in crime rates in affected areas. \\\cline{3-4}
& & Avoid Strategy & I believe that prioritizing the expansion of civil liberties may not always lead to the best outcomes for society as a whole. Sometimes, certain restrictions are necessary to maintain order and stability. By focusing on other aspects of society, we can ensure a more balanced and harmonious environment for everyone. \\
\hline

\multirow{4}{*}{Emotional} 
& \multirow{2}{*}{We should not support the widespread adoption of organic food.} 
& Use Strategy & Let's not forget the heart-wrenching reality of organic farming. Imagine a world where organic farming is the norm, but the yield is significantly lower. Picture families, single parents, and the elderly, struggling to afford the higher prices of organic food. Imagine the fear and desperation of a mother, unable to provide for her children because the food is just too expensive. The dream of organic food for all is beautiful, but the potential nightmares it could bring. \\\cline{3-4}
& & Avoid Strategy & However, I believe that the widespread adoption of organic food may not be the most practical solution for addressing sustainability in food production. Conventional farming methods can also implement techniques to improve soil health and reduce pollution without solely relying on organic practices. Additionally, the lower yields associated with organic farming could potentially lead to food shortages and higher prices, which may not be feasible for meeting the global food demand. \\
\cline{2-4}

& \multirow{2}{*}{We should not allow hunting altogether.} 
& Use Strategy & Imagine a deer, wasting away, its ribs showing through its thinning coat, struggling for every meager mouthful of food in the dead of winter. Consider the spread of disease that could wipe out entire herds, entire species. It’s a heart-wrenching image, isn't it? \\\cline{3-4}
& & Avoid Strategy & While regulated hunting might be a solution to overpopulation problems, it's important to consider its potential risks and ethical issues. The impact of hunting on individual animals and species populations is significant, and non-lethal methods and ecosystem management practices could be a more sustainable approach to these challenges. \\
\hline

\multirow{4}{*}{Moral} 
& \multirow{2}{*}{We should not implement censorship.} 
& Use Strategy & I believe that censorship should not be implemented, even in the name of finding a middle ground between safety and freedom of expression. While it is crucial to protect individuals from harm, censorship can lead to the suppression of diverse perspectives and ideas. It is essential to trust in the ability of individuals to critically evaluate information and make their own decisions. Upholding the fundamental right to free speech is crucial for promoting a society that values open dialogue, creativity, and progress. \\\cline{3-4}
& & Avoid Strategy & While the idea of censorship might seem like an efficient solution to control misinformation, its effectiveness can be questionable. The risk of unintentionally censoring valid content or diverse opinions is high. Instead, it would be more practical to invest in enhancing media literacy, developing critical thinking skills, and establishing reliable fact-checking systems. These can help individuals make informed judgments about the credibility of the information they come across. \\
\cline{2-4}

& \multirow{2}{*}{We should support the Fat Acceptance Movement.} 
& Use Strategy & While promoting healthy choices is important, it's also crucial to recognize that individuals come in all shapes and sizes. The Fat Acceptance Movement focuses on ending weight-based discrimination and promoting respect for people of all body types. By supporting this movement, we are advocating for the rights and dignity of individuals who have long been marginalized and stigmatized based on their weight. It's about promoting inclusivity, understanding, and compassion for everyone, regardless of their size. \\\cline{3-4}
& & Avoid Strategy & Let's consider this. The Fat Acceptance Movement is not encouraging unhealthy habits. Instead, it's highlighting that health isn't solely determined by weight. It's suggesting that we must consider various factors when judging healthiness, not just body size. This approach proposes a more comprehensive view of health, indicating that people can be healthy at different sizes. \\
\hline

\end{tabular}
}
\caption{Examples of generated utterances across four persuasion strategies (Causal, Empirical, Emotional, Moral) under different instruction conditions (Use vs. Avoid). Each entry includes the topic and the model’s response based on the strategic prompt.}
\label{tab:utterance_examples}
\end{table}

\clearpage

\twocolumn
\section{GPT Annotation Scores}

To evaluate the effectiveness of the rhetorical constraints described in Section \ref{sec:dialogue}, we conducted a comprehensive analysis of how well the generated utterances reflected the intended persuasive strategies. Specifically, we examined the distributions of LLM-generated scores for utterances that were explicitly conditioned to either use (positive) or avoid (negative) each of the four strategies: causal, empirical, moral, and emotional. These scores were produced by five persona-conditioned GPT-4o annotators, each independently rating the presence of each rhetorical strategy on a five-point Likert scale.

Figure \ref{fig:gptscores} visualizes the resulting distributions, showing that utterances generated under the "use" condition consistently received higher scores than those under the "avoid" condition for every strategy.

To quantify this effect more precisely, we calculated the Spearman rank correlation between the binary assignment (use vs.\ avoid) and the corresponding averaged LLM strategy scores. As shown in Table~\ref{tab:spearman_gptlabels} , we found strong positive correlations for all four rhetorical strategies: $\rho = 0.863$ for \textit{moral}, $\rho = 0.785$ for \textit{emotional}, $\rho = 0.812$ for \textit{causal}, and $\rho = 0.805$ for \textit{empirical}. These results demonstrate that the generation system effectively controlled for rhetorical style, and that the use of rhetorical constraints yielded outputs that aligned closely with the intended persuasive strategies, as judged by independently simulated LLM annotators.



\section{Human annotation}
\label{human-annotation}
For each argument, the participant sees four separate prompts along with a reminder of the definition of the strategy at question:
\begin{enumerate}
    \item Is this argument empirical?
    
Here again is the definition of empirical:

An empirical argument relies on evidence such as statistics, examples, illustrations, anecdotes, and/or citations to sources that support the argument.
    \item Is this argument causal?

Here again is the definition of logical:

A causal argument relies on cause-and-effect reasoning to explain or predict the positive or negative consequences of an action that are measurable or observable, with or without evidence.
    \item Is this argument emotional?

Here again is the definition of emotional:

An emotional argument relies on impassioned, arousing, or provocative language to express or evoke feelings (such as frustration, fear, hope, joy, desire, sadness, hurt, and/or surprise), rather than relying on rational or moral appeals.
    \item Is this argument moral?

Here again is the definition of moral:

A moral argument relies on concepts of right and wrong, justice, virtue, duty, or the greater good in order to persuade others about the ethical merit of an action.
\end{enumerate}

We provide five options to choose for each prompt:
\begin{enumerate} 
    \item Definitely not 
    \item Probably not 
    \item Might or might not 
    \item Probably yes 
    \item Definitely yes 
\end{enumerate}

\section{Annotator Training Procedure}
\label{sec:trainingprocedure}
Before human participants begin their annotation task, they are asked to take a quiz. In the quiz, they are introduced to the definition and two examples of every strategy, presented with two arguments, one using the strategy and the other not, and asked to label them as either using the strategy or not. The examples and quiz arguments are drawn from the same samples as the LLM-based labeling.

If a participant labels every quiz argument correctly, they proceed to start the annotation task. Otherwise, they are redirected to a second round of quiz, which repeats the procedure above, with different examples and quiz arguments. 

\begin{table*}[!t]
\centering
\resizebox{\textwidth}{!}{
\begin{tabular}{p{2.5cm}p{4.5cm}p{4cm}p{11cm}}
\hline
\textbf{Strategy Type} & \textbf{Evaluation Dataset} & \textbf{Relevant Label} & \textbf{Label Definition} \\
\hline
\textbf{Causal} & Fallicious Argument Classification \citep{ijcai2022p575} & Slippery Slope & {\small It suggests that an unlikely exaggerated outcome may follow an act. The intermediate premises are usually omitted and a starting premise is usually used as the first step leading to an exaggerated claim.} \\
\cline{3-4}
& & False Cause & {\small The misinterpretation of the correlation of two events for causation \citep{Walton_1987}.} \\
\cline{2-4}
& Persuasion For Good \citep{wang-etal-2019-persuasion} & Logical Appeal & {\small The use of reasoning and evidence to convince others. For instance, a persuader can convince a persuadee that the donation will make a tangible positive impact for children using reasons and facts.} \\
\hline
\textbf{Empirical} & Persuasion For Good \citep{wang-etal-2019-persuasion} & Credibility & {\small Use of credentials and citing organizational impacts to establish credibility and earn the persuadee’s trust. The information usually comes from an objective source (e.g., the organization’s website or other well-established websites).} \\
\cline{2-4}
& Good Faith Textual Requests \citep{Chen_Yang_2021} & Evidence & {\small Providing concrete facts or evidence for the narrative or request.} \\
\cline{2-4}
& Fallicious Argument Classification \citep{ijcai2022p575} & Appeal to Authority & {\small When the arguer mentions the name of an authority or a group of people who agreed with her claim either without providing any relevant evidence, or by mentioning popular non-experts, or the acceptance of the claim by the majority.} \\
\hline
\textbf{Emotional} & Fallicious Argument Classification \citep{ijcai2022p575} & Appeal to Emotion & {\small The unessential loading of the argument with emotional language to exploit the audience emotional instinct.} \\
\cline{2-4}
& Persuasion For Good \citep{wang-etal-2019-persuasion} & Personal Story & {\small Using narrative exemplars to illustrate someone’s donation experiences or the beneficiaries’ positive outcomes, which can motivate others to follow the actions.} \\
\hline
\textbf{Moral} & Moral Emotion Dataset \citep{kim-etal-2024-moral} & Moral Emotion (Existence of any of the four emotional strategy labels by majority vote) & 
\parbox[t]{11cm}{
    \small \textbf{Other-condemning}: Condemn others (e.g., anger, contempt, disgust)\\
    \small \textbf{Other-praising}: Praise others (e.g., admiration, gratitude, awe)\\
    \small \textbf{Other-suffering}: Empathy for the suffering of others (e.g., compassion, sympathy)\\
    \small \textbf{Self-conscious}: Negatively evaluate oneself (e.g., shame, guilt, embarrassment)
} \\
\hline
\end{tabular}
}
\caption{\textbf{Label Definitions for External Evaluation.} This table describes the relevant rhetorical or logical labels associated with each strategy type and dataset used in external validation. Due to dataset variation, only approximate matches to our strategy dimensions are used.}
\label{tab:label_definitions}
\end{table*}

We provide two chances for each participant to label all arguments in a quiz correctly. If they fail to do so by the end of the second quiz, they are automatically directed to exit the survey.

\newpage
\section{Inter-rater Consistency: LLM}
\label{sec:irr}

\begin{table}[ht]
\centering
\small
\resizebox{\columnwidth}{!}{
\begin{tabular}{lccc}
\toprule
& \multicolumn{3}{c}{\textbf{Classification Scheme}} \\
\cline{2-4}
\textbf{Rhetorical Strategy} & \textbf{Five-Class} & \textbf{Three-Class} & \textbf{Two-Class} \\
&\textbf{(Original Scheme)}&& \\
\midrule
Causal     & 0.458& 0.665 & 0.749\\
Empirical   & 0.595 & 0.672 & 0.793\\
Moral       & 0.566 &  0.692& 0.829\\
Emotional  & 0.546 &  0.691 & 0.822\\
\midrule
\textbf{Average}  & 0.541 & 0.680 & 0.798 \\
\bottomrule
\end{tabular}
}
\caption{\textbf{Inter-rater consistency (Cohen’s Kappa) of LLM annotators under the five-class, three-class and two-class classification scheme.} LLMs consistently demonstrate substantially higher internal reliability than human annotators.}
\label{tab:internal_consistency}
\end{table}

\section{Presidential Debate Human Validation}
\label{sec:human_valid_president}

\begin{table}[ht]
\centering
\begin{tabular}{lc}
\hline
\textbf{Strategy} & \textbf{Spearman's $\rho$ (Model vs. Human)} \\
\hline
Causal     & 0.618 \\
Empirical  & 0.614 \\
Moral      & 0.618 \\
Emotional  & 0.567 \\
\hline
\end{tabular}
\caption{Comparison of RoBERTa model predictions with human-annotated strategy scores on the presidential debate dataset. The values indicate Spearman’s rank correlation ($\rho$) between model predictions and average human annotations.}
\label{tab:roberta_human_corr}
\end{table}


\section{Label Definitions for External Validation Datasets}
\label{sec:external_def}

The label definitions from external datasets used in our external validity tests, which are relevant to the rhetorical typology in our experiment, are presented in Table~\ref{tab:label_definitions}.

\twocolumn

\section{Persuasiveness Score Datasets}
\label{sec:datapers}

\vspace{1ex}

\textbf{ConvArg} \citep{habernal2016makes}: The ConvArg dataset contains 9,111 argument pairs from the online debate platforms CreateDebate and ConvinceMe. Each argument pair is annotated via crowdsourcing, where human annotators indicate which argument is more convincing through a binary judgment and justify their choice by selecting from a set of predefined reasons, including strength of reasoning, emotional appeal, relevance to the topic, and language quality. We compute the persuasiveness score for each argument based on pairwise argument quality, using the PageRank \citep{simpson2018finding} or winning rate as $Score = \frac{\#Win}{\#Win + \#Loss}$
where \(\#Win\) denotes the number of times an argument is labeled more persuasive, and \(\#Loss\) the number of times it is deemed less persuasive. This method mirrors the one proposed by \citet{gretz2020large}.

\vspace{1ex}

\textbf{IBM$\_$30k} \citet{gretz2020large}:
The IBM-Rank-30k dataset contains 30,497 crowd-sourced arguments on 71 controversial topics, collected via the Figure Eight platform. For each topic, annotators were asked to write two short arguments—one supporting and one opposing the topic—as if preparing for a public speech. Each argument was then evaluated by 10 annotators, who were asked whether they would recommend the argument to a friend preparing a speech, regardless of their personal stance on the issue.  To derive a continuous quality score from these binary responses, the dataset employs a Weighted Average scoring function, which adjusts each annotator’s influence based on their annotator reliability score—a measure of how consistently the annotator agrees with others across previous shared tasks.

\vspace{1ex}

\textbf{IBM$\_$5.3k} \citep{toledo2019automatic}: IBM$\_$5.3k consists of 5.3k arguments selected from the UKPConvArg database \citep{habernal_second}, originally sourced from the Reddit CMV forum. Each argument has two types of labels: an individual argument quality label (absolute) and a relative argument-pair label (relative). For the absolute label, annotators are asked a binary yes/no question about whether they would recommend a friend preparing a speech supporting or contesting the topic to use the argument. The quality of each individual argument is a real-valued score between 0 and 1, defined by the fraction of ‘yes’ responses.

For the relative label, annotators are presented with a pair of arguments that take the same stance on a topic and are asked which of the two would be preferred by most people to support or contest the topic. The final dataset consists of 5.3k arguments, each selected based on high individual quality ratings, with an average of 11.4 valid annotations per argument.

\vspace{1ex}

\textbf{IAC} \citep{iac}:
The dataset comprises 109,074 sentences covering four debate topics—gay marriage, gun control, the death penalty, and evolution—sourced from the IAC corpus \citet{walker-etal-2012-corpus} and CreateDebate.com. Each sentence was annotated by seven Amazon Mechanical Turk workers with approval ratings above $95\%$. Annotations include a binary label indicating whether the sentence expresses an argument, as well as a continuous argument score ranging from 0 (difficult to interpret) to 1 (easy to interpret).

\vspace{1ex}

\textbf{IDEA} \citep{IDEA}: The IDEA dataset consists of 165 debates obtained from the International Debate Education Association website. Each debate includes a motion that expresses a stance on a topic, along with an average of 7.3 arguments either supporting or opposing the motion. Each argument contains a one-sentence assertion of its stance and a justification explaining that stance. Two native English speakers annotated each argument with a persuasiveness score from 1 to 6, along with five types of errors that may have undermined its persuasiveness: grammar errors, lack of objectivity, inadequate support, unclear assertion, and unclear justification.

\begin{table*}[!t]

\centering
\resizebox{\textwidth}{!}{
\begin{tabular}{l|cc|cc|cc|cc|cc}
\toprule
& \multicolumn{2}{c|}{ConvArg (1038)} & \multicolumn{2}{c|}{IBM-30k (30497)} & \multicolumn{2}{c|}{IBM-5.3k (5298)} & \multicolumn{2}{c|}{IAC (4939)} & \multicolumn{2}{c}{IDEA (1205)} \\
\midrule
\midrule
& \textbf{Spearman's} $\mathbf{\rho}$$\uparrow$& \textbf{RMSE} $\downarrow$& \textbf{Spearman's} $\mathbf{\rho}$ $\uparrow$& \textbf{RMSE} $\downarrow$& \textbf{Spearman's} $\mathbf{\rho}$ $\uparrow$& \textbf{RMSE} $\downarrow$& \textbf{Spearman's} $\mathbf{\rho}$ $\uparrow$& \textbf{RMSE} $\downarrow$& \textbf{Spearman's} $\mathbf{\rho}$ $\uparrow$& \textbf{RMSE} $\downarrow$\\
\midrule
\textbf{Within Dataset - Vanilla} & 0.647 (0.012) & 0.265 (0.004) & 0.502 (0.004) & 0.176 (0.004) & 0.456 (0.010) & 0.204 (0.004) & 0.670 (0.000) & 0.188 (0.008) & 0.263 (0.021) & 0.280 (0.007) \\
\textbf{Within Dataset - Strategy} & \textbf{0.680} (0.009) & \textbf{0.255} (0.003) & \textbf{0.516} (0.005) & \textbf{0.167} (0.003) & \textbf{0.478} (0.009) & \textbf{0.188} (0.004) & \textbf{0.678} (0.003) & \textbf{0.171} (0.005) & \textbf{0.337} (0.036) & \textbf{0.264} (0.007) \\
$\boldsymbol{\Delta}$ & +0.033 & -0.010 & +0.014 & -0.009 & +0.022 & -0.016 & +0.008 & -0.017 & +0.074 & -0.016 \\
\midrule
\midrule
\textbf{Cross Dataset - Vanilla} & 0.300 (0.018) & 0.335 (0.003) & 0.290 (0.005) & 0.247 (0.019) & 0.380 (0.005) & 0.345 (0.004) & 0.349 (0.003) & 0.283 (0.004) & 0.052 (0.010) & 0.396 (0.005) \\
\textbf{Cross Dataset - Strategy} & \textbf{0.341} (0.016) & \textbf{0.326} (0.001) & \textbf{0.309} (0.009) & \textbf{0.218} (0.012) & \textbf{0.400} (0.005) & \textbf{0.335} (0.004) & \textbf{0.389} (0.014) & \textbf{0.257} (0.008) & \textbf{0.053} (0.009) & \textbf{0.395} (0.004) \\
$\boldsymbol{\Delta}$ & +0.041 & -0.009 & +0.019 & -0.029 & +0.020 & -0.010 & +0.040 & -0.026 & +0.001 & -0.001 \\
\bottomrule

\end{tabular}
}
\caption{Persuasiveness score performance for the vanilla model and the model augmented with rhetorical strategies, evaluated within and across datasets. The table reports the mean and standard deviation of performance across three fine-tuning runs. $\Delta$ rows report the difference between strategy-enhanced and vanilla models (higher $\rho$, lower RMSE is better).}
\label{tab:persuassiveness-appendix}
\end{table*}

\begin{table*}[!t]
\centering
\resizebox{\textwidth}{!}{
\begin{tabular}{l|cc|cc|cc|cc|cc}
\toprule
& \multicolumn{2}{c|}{ConvArg (1038)} & \multicolumn{2}{c|}{IBM-30k (30497)} & \multicolumn{2}{c|}{IBM-5.3k (5298)} & \multicolumn{2}{c|}{IAC (4939)} & \multicolumn{2}{c}{IDEA (1205)} \\
\midrule
\midrule
& \textbf{Test Set Size} & \textbf{Mean SE} & \textbf{Test Set Size} & \textbf{Mean SE} & \textbf{Test Set Size} & \textbf{Mean SE} & \textbf{Test Set Size} & \textbf{Mean SE} & \textbf{Test Set Size} & \textbf{Mean SE} \\
\midrule
\textbf{Within Dataset - Vanilla} & 104 & 0.019 & 3050 & 0.002 & 530 & 0.005 & 494 & 0.005  & 120 &  0.013\\
\textbf{Within Dataset - Strategy} & 104 & 0.014 & 3050 & 0.002 & 530 & 0.004 & 494 & 0.005 & 120 & 0.009 \\
\midrule
\midrule
\textbf{Cross Dataset - Vanilla} & 1038 &  0.007 & 30497 & 0.000 & 5298 & 0.003 & 4939 & 0.003 & 1205 &  0.008 \\
\textbf{Cross Dataset - Strategy} & 1038 & 0.007  & 30497 & 0.000 & 5298 & 0.002 & 4939 &  0.002 & 1205 & 0.008\\
\bottomrule
\end{tabular}
}
\caption{Test set size for each dataset and the average standard error (Mean SE) of the persuasiveness prediction model over three fine-tuning runs.}
\label{tab:persuassiveness_testsize}
\end{table*}

\section{Persuasiveness Prediction Performance Details}
We evaluated persuasiveness prediction performance using two complementary metrics: Spearman correlation between predicted and ground-truth persuasiveness scores, and Root Mean Squared Error (RMSE). To assess the contribution of rhetorical features, we compared model performance under two conditions—with and without the inclusion of predicted strategy scores. As shown in \Cref{tab:persuassiveness-appendix} reports small but consistent improvements in predicting persuasiveness when incorporating rhetorical strategy. Within-domain, the strategy features increased the correlation with ground-truth persuasiveness scores by 0.03 and reduced RMSE by 0.014. These effect sizes are small but not negligible. For each dataset, the differences are statistically significant, with sample sizes ranging from hundreds to thousands of observations. The improvements represent a relative $8.40\%$ increase in the magnitude of the correlations and a relative $6.30\%$ decrease in RMSE, indicating better alignment with human judgments. In the more challenging cross-domain setting, we observe a relative $7.77\%$ increase in correlation and a relative $6.16\%$ reduction in RMSE. These results suggest that incorporating rhetorical strategies improves prediction not only within individual domains but also in previously unseen topical contexts.

\section{Individual Debate Level Strategy Comparisons}
\label{sec:debate_individual}

\begin{table}[H]
\centering
\resizebox{\columnwidth}{!}{

\begin{tabular}{llrrrr}

\toprule
\textbf{Year} & \textbf{Candidates}& \textbf{Causal} & \textbf{Empirical} & \textbf{Emotional} & \textbf{Moral} \\
\midrule
1960 & Kennedy (D) vs. Nixon (R)& 0.005 & 0.031 & -0.041 & -0.041 \\
1976 & Carter (D) vs. Gerald Ford (R)& 0.002 & 0.014 & 0.090$^{***}$ & 0.008 \\
1980 & Reagan (R) vs. Jimmy Carter (D)& 0.055& 0.010& -0.074$^{*}$& 0.036\\
1984 & Reagan (R) vs. Mondale (D)& 0.081$^{**}$& -0.011& 0.087$^{***}$& 0.087$^{*}$\\
1988 & H. W. Bush (R) vs. Dukakis (D)& 0.041& 0.000& -0.008& 0.041\\
1992 & B. Clinton (D) vs. H. W. Bush (R)& 0.078$^{**}$ & 0.193$^{***}$ & -0.019 & 0.022 \\
1996 & B. Clinton (D) vs. Dole (R)& 0.067$^{**}$ & 0.067$^{*}$ & -0.036$^{*}$ & 0.056$^{*}$ \\
2000 & G. W. Bush (R) vs. Gore (D)& -0.006& 0.058$^{**}$& -0.033$^{*}$& -0.043$^{*}$\\
2004 & G. W. Bush (R) vs. John Kerry (D)& -0.022& 0.021& 0.023& -0.011\\
2008 & Obama (D) vs. McCain (R)& 0.025 & -0.001 & -0.052$^{***}$ & -0.004 \\
2012 & Obama (D) vs. Romney (R)& 0.047$^{*}$ & 0.016 & 0.000 & 0.017 \\
2016 & Trump (R) vs. H. Clinton (D)& 0.030$^{*}$& 0.047$^{***}$& -0.049$^{***}$& 0.050$^{***}$\\
2020 & Biden (D) vs. Trump (R)& 0.006 & 0.016 & -0.024$^{*}$ & 0.045$^{***}$ \\
\bottomrule
\end{tabular}
}
\vspace{0.5em}
\begin{minipage}{\columnwidth}
\raggedright
\textit{Note.} Entries reflect the difference in strategy score (Democrat minus Republican) averaged over all utterances of each strategy type.\\
$^{*} p < 0.05$, $^{**} p < 0.01$, $^{***} p < 0.001$
\end{minipage}
\caption{\textbf{Partisan Differences in Rhetorical Strategy Between U.S. Presidential Debate Candidates.} Differences are measured as the Democrat's average over scores for each utterance, minus the Republican's average, broken down by four rhetorical strategies:.}
\label{tab:debate_all_strategy_types}
\end{table}
\end{document}